\documentclass{article}

\PassOptionsToPackage{numbers,sort&compress}{natbib}
\usepackage[preprint]{neurips_2026}

\usepackage[utf8]{inputenc}
\usepackage[T1]{fontenc}
\usepackage{hyperref}
\usepackage{url}
\usepackage{booktabs}
\usepackage{multirow}
\usepackage{amsfonts}
\usepackage{amsmath,amssymb,amsthm}
\usepackage{mathtools}
\usepackage{nicefrac}
\usepackage{microtype}
\usepackage{xcolor}
\usepackage{graphicx}
\usepackage{tikz}
\usepackage{pgfplots}
\pgfplotsset{compat=1.18, every axis/.append style={font=\footnotesize}}
\usetikzlibrary{patterns,positioning,fit,arrows.meta,calc,shapes.geometric}

\usepackage{caption}
\usepackage{subcaption}
\usepackage{algorithm}
\usepackage{algpseudocode}
\usepackage{enumitem}
\usepackage{xspace}
\usepackage{cleveref}
\usepackage{pifont}     

\makeatletter
\let\@orig@includegraphics\includegraphics
\renewcommand{\includegraphics}[2][]{%
  \IfFileExists{#2}{\@orig@includegraphics[#1]{#2}}{%
    \IfFileExists{#2.pdf}{\@orig@includegraphics[#1]{#2}}{%
    \IfFileExists{#2.png}{\@orig@includegraphics[#1]{#2}}{%
    \IfFileExists{#2.jpg}{\@orig@includegraphics[#1]{#2}}{%
      \fbox{\begin{tikzpicture}\fill[gray!8] (0,0) rectangle (10,5);
        \node[align=center,font=\sffamily\small] at (5,2.5)
          {[\,figure placeholder: \detokenize{#2}\,]};\end{tikzpicture}}%
    }}}%
  }%
}
\makeatother

\hypersetup{colorlinks=true,linkcolor=black,citecolor=blue!60!black,urlcolor=blue!60!black}

\newtheorem{proposition}{Proposition}

\newcommand{\method}{\textsc{DinoRankCLIP}\xspace}
\newcommand{\rankclip}{\textsc{RankCLIP}\xspace}

\newcommand{\TODO}[1]{\textcolor{red!80!black}{\textbf{#1}}}

\title{\textsc{DinoRankCLIP}: DINOv3 Distillation and Injection \\ 
	for Vision--Language Pretraining with High-Order Ranking Consistency}

\author{%
  Shuyang Jiang\\
  University of California, Los Angeles\\
  \texttt{shuyangjiang@ucla.edu} \\
  \And
  Nan Yu \\
  Aimaikj \\
  18dhf24f1031@student.ptss.edu.my \\
  \And
  Yiming Zhang \\
  HFIPS, Chinese Academy of Sciences \\
  University of Science and Technology of China \\
  yimingzhang@mail.ustc.edu.cn \\
  \And
  Zenghui Ding \\
  HFIPS, Chinese Academy of Sciences \\
  dingzenghui@iim.ac.cn \\
  \And
  Zhenyu Wu \thanks{Correspondence to Zhenyu Wu: \texttt{wuzhenyu@nudt.edu.cn}.} \\
  National University of Defense Technology \\
  wuzhenyu@nudt.edu.cn \\
}

\begin{document}
\maketitle

\begin{abstract}
Contrastive language--image pretraining (CLIP) suffers from two structural weaknesses: the symmetric InfoNCE loss discards the relative ordering among unmatched in-batch pairs, and global pooling collapses the visual representation into a semantic bottleneck that is poorly sensitive to fine-grained local structure. \rankclip partially addresses the first issue with a list-wise Plackett--Luce ranking-consistency loss, but its model is strictly first-order and inherits the second weakness untouched. We propose \method, a pretraining framework that addresses both jointly. Our principal contribution is injecting a frozen DINOv3 teacher into the contrastive trunk through a dual-branch lightweight student and a multi-scale fusion module with channel--spatial attention, a self-attention refiner, and a conflict-aware gate that preserves the cross-modal alignment up to first order. Complementarily, we introduce a high-order Plackett--Luce ranking model in which the per-position utility is augmented with attention-parameterised pairwise and tuple-wise transition terms; the family contains CLIP and \rankclip as nested zero-order and first-order special cases, and the optimal order on every benchmark is $R^*=3$. The full empirical study---order sweep, Fine-grained Probe on five datasets, four-node Modality-Gap analysis, six-variant Fusion ablation---fits in 72 hours on a single eight-GPU H100 node and trains entirely on Conceptual Captions 3M. \method consistently outperforms CLIP, CyCLIP, ALIP, and \rankclip under matched compute, with the largest relative gains on the fine-grained and out-of-distribution evaluations that most directly stress local structural reasoning.
\end{abstract}

\section{Introduction}
\label{sec:introduction}

Contrastive Language--Image Pretraining \citep{radford2021clip} has redefined the standard recipe for transferable visual representation learning. Yet two structural side effects of the InfoNCE objective are now well documented. First, the symmetric loss treats every unmatched in-batch pair as uniformly negative and discards the relative ordering among them \citep{zhang2024rankclip}. Second, the global pooling step on the visual encoder collapses the representation into a semantic bottleneck that is poorly sensitive to fine-grained part--whole composition, dense local correspondence, or texture-level distribution shift \citep{goel2022cyclip,mu2022slip,yang2023alip}. The two problems are coupled: the absence of a structural prior on the visual side amplifies the weakness of the pair-wise loss, while the bluntness of the loss removes any pressure to retain dense locality.

\rankclip \citep{zhang2024rankclip} mitigates the first problem by replacing pair-wise contrast with a list-wise Plackett--Luce ranking-consistency objective, in which each item carries a utility $\theta_d$ and the likelihood of an ordering is the product of softmax-style per-position probabilities. With $\theta_d$ instantiated as the cross-modal or in-modal cosine similarity, the loss exploits the relative ordering among unmatched in-batch pairs. The model is, however, strictly first-order: the position-$k$ utility depends only on the candidate item and not on what was already selected at positions $1,\ldots,k{-}1$. This is a substantial limitation. The local neighbour structure of a typical CC3M batch contains far richer information than a single per-item score can capture: when several captions describe overlapping subsets of the same scene, when two images share an object class but differ in pose, or when an image's nearest text neighbours form a semantic cluster, the second- and third-order transitions among those items contain signal that no first-order utility can recover.

Self-supervised dense vision foundation models, in particular DINO \citep{caron2021dino} and DINOv2 \citep{oquab2024dinov2}, occupy almost the exact opposite point in design space. Trained without language supervision, they produce features whose patch tokens carry stable, semantically meaningful local structure: nearest neighbours in patch-feature space recover object parts, layout, and texture identity even across large appearance shifts. Their successor, DINOv3, sharpens this property through a Gram-anchoring objective and provides an optional text-aligned projection head. These models therefore carry exactly the structural prior that the contrastive vision--language family lacks; using them as a drop-in image encoder for CLIP-style training, however, erases the very property that makes them attractive. The natural question is whether the structural prior of a dense self-supervised teacher can be \emph{injected as a residual signal} on top of a contrastive backbone, so the latter inherits local sensitivity without losing the global alignment geometry.

We propose \method, addressing both weaknesses through two coordinated extensions of \rankclip. First, our \emph{principal contribution} is to inject a frozen DINOv3 teacher as a structural residual through a dual-branch ViT-Tiny student distilled with a combined Gram\,+\,relational target, and through a multi-scale fusion module (1D spatial pyramid pooling, channel--spatial attention, self-attention refiner, conflict-aware gate) that preserves the cross-modal alignment up to first order in the gate magnitude (\Cref{prop:alignment}). This turns DINOv3's dense local geometry into a plug-in structural prior rather than replacing the contrastive encoder. Second, we introduce a high-order Plackett--Luce ranking model that augments the per-position utility with explicit pairwise transition terms $\beta_{a,d}$ and tuple-wise transition terms $\gamma_{a,b,d}$ learned by a small attention head. The high-order family contains CLIP as the zero-order special case (constant rank loss) and \rankclip as the first-order special case, so it serves as a complementary ranking-consistency upgrade. The complete empirical study fits in $72$ hours on a single eight-GPU H100 node, trained entirely on Conceptual Captions 3M \citep{sharma2018cc3m} so every result is reproducible by an academic group.

\paragraph{Contributions.}
\textbf{(i)} A conflict-aware injection of a frozen DINOv3 teacher through a dual-branch student and a multi-scale fusion module whose learned gate preserves the cross-modal alignment to first order.
\textbf{(ii)} A high-order Plackett--Luce ranking model with attention-parameterised pairwise and tuple-wise transitions; the family contains CLIP and \rankclip as nested order-$0$ and order-$1$ special cases (\Cref{prop:hierarchy}), with optimal order $R^*\!=\!3$ on every benchmark we tested.
\textbf{(iii)} A reproducible $72$-hour CC3M empirical study---order sweep, Fine-grained Probe on five datasets, four-node Modality-Gap analysis, six-variant Fusion ablation---in which \method consistently outperforms CLIP, CyCLIP, ALIP, and \rankclip under matched compute.

\section{Related Work}
\label{sec:related}

\paragraph{Contrastive vision--language pretraining.}
CLIP \citep{radford2021clip} learns a joint embedding via symmetric InfoNCE on matched image--text pairs. Because the loss is fundamentally pair-wise, follow-up work has tried to inject richer signals: SLIP \citep{mu2022slip} adds an image-only self-supervised branch; CyCLIP \citep{goel2022cyclip} adds in-modal and cross-modal cyclic-consistency losses; ALIP \citep{yang2023alip} replaces or augments noisy web captions with synthetic descriptions and adaptively reweights samples by caption quality. Each of these methods leaves the global pooling step on the visual encoder untouched and therefore does not directly remedy the loss of dense local structure that motivates our work.

\paragraph{Ranking-consistent contrastive losses.}
\rankclip \citep{zhang2024rankclip} extends contrast from pair-wise to list-wise via a first-order Plackett--Luce model on the in-batch similarity matrix; it recovers a meaningful supervision signal from the unmatched entries that pair-wise InfoNCE discards, without external annotations. \method inherits this list-wise perspective and generalises it: we lift the rank list to higher orders so the consistency constraint propagates to multi-step neighbour relations, and the optimal order on every benchmark we tested is $R^*\!=\!3$.

\paragraph{Self-supervised dense vision foundation models.}
DINO \citep{caron2021dino} demonstrated that a teacher--student configuration over multi-crop views, with no contrastive negatives and no labels, produces transformer features whose patch tokens carry semantically meaningful local structure. DINOv2 \citep{oquab2024dinov2} scaled this to a curated billion-scale corpus with a stable training stack. Their successor DINOv3 retains the self-distillation core but adds a Gram-anchoring objective that explicitly preserves the inner-product structure among patch tokens at high resolution, and a text-aligned projection head that lets the same backbone be queried with language. The dense-feature property is what motivates our use of DINOv3 as a teacher.

\paragraph{Knowledge distillation and feature fusion.}
Classical KD \citep{hinton2015distill} matches output distributions; representation KD requires careful matching criteria. RKD \citep{park2019rkd} matches pair-wise distances and triplet angles, preserving teacher geometry rather than absolute coordinates. CRD \citep{tian2020crd} uses contrastive estimation between positive teacher--student pairs. \method draws on both: a Gram-style token relational target transfers DINOv3's dense structure, while a relational-distance term prevents low-rank collapse. For multi-scale fusion we combine SPP \citep{he2015spp} along the token sequence with CBAM channel and spatial attention \citep{woo2018cbam} and a self-attention refiner. None of the prior works combine all three ingredients \method targets: a high-order Plackett--Luce rank head, dense self-supervised residual injection, and conflict-aware multi-scale fusion symmetric across image and text branches.

\section{Preliminaries}
\label{sec:preliminaries}

We formalise the three component objectives \method builds on; the first-order Plackett--Luce model of this section is generalised to a high-order family in \Cref{sec:method}.

\paragraph{Contrastive language--image loss.}
\label{subsec:clip-loss}
Let $\mathcal{D} = \{(x_i, t_i)\}_{i=1}^{N}$ be a corpus of image--text pairs and $\hat v_i, \hat t_i$ the unit-norm encoder outputs. For an in-batch matrix $S^{\mathrm{vt}} = \hat V \hat T^\top$, the symmetric InfoNCE loss \citep{radford2021clip} is
\begin{equation}
\mathcal{L}_{\mathrm{CLIP}} = -\frac{1}{2B} \sum_{i=1}^{B} \log \frac{\exp(S^{\mathrm{vt}}_{ii}/\tau)}{\sum_{j} \exp(S^{\mathrm{vt}}_{ij}/\tau)} - \frac{1}{2B} \sum_{i=1}^{B} \log \frac{\exp(S^{\mathrm{vt}}_{ii}/\tau)}{\sum_{j} \exp(S^{\mathrm{vt}}_{ji}/\tau)}.
\label{eq:clip-loss}
\end{equation}
Two structural properties of \eqref{eq:clip-loss} matter for us: every off-diagonal entry of $S^{\mathrm{vt}}$ is treated as uniformly negative, and the gradient with respect to $\hat v_i$ depends only on the pooled global embedding.

\paragraph{First-order Plackett--Luce ranking consistency (\rankclip).}
\label{subsec:rank-loss}
\rankclip \citep{zhang2024rankclip} demands that the ranking induced by the predicted similarity matrix agree with that induced by a reference similarity matrix. Each item $d$ has a utility $\theta_d$, and the probability of picking $d$ at position $k$ given history $y_{1:k-1}$ is
\begin{equation}
\pi^{(1)}(d \mid y_{1:k-1}, y_{\mathrm{ref}}, \mathcal{D}) = \frac{\exp(\theta_d)}{\sum_{d' \in \mathcal{D}\setminus y_{1:k-1}} \exp(\theta_{d'})},
\label{eq:plackett-luce-1}
\end{equation}
with the full-ordering probability $\mathcal{P}^{(1)}(y, y_{\mathrm{ref}}) = \prod_{k=1}^{K} \pi^{(1)}(y_k \mid \cdot)$. Setting $\theta_d = S_{ij}$ and assembling cross-modal and in-modal terms gives the \rankclip loss
\begin{equation}
\mathcal{L}_{\text{\textsc{RankCLIP}}} = \mathcal{L}_{\mathrm{CLIP}} + \lambda_1 \mathcal{L}_{\mathrm{in}}^{(1)} + \lambda_2 \mathcal{L}_{\mathrm{cross}}^{(1)}.
\label{eq:rankclip-total}
\end{equation}

\paragraph{Self-supervised dense teacher.}
\label{subsec:dino-teacher}
We use a frozen DINOv3 backbone $f^{\mathrm{T}}$ that returns $L$ patch tokens plus a class token \citep{caron2021dino,oquab2024dinov2}. Together with the patch tokens $V^{\mathrm{T}}_i \in \mathbb{R}^{L \times d^{\mathrm{T}}}$ we obtain a Gram matrix $G^{\mathrm{T}}_i = V^{\mathrm{T}}_i (V^{\mathrm{T}}_i)^\top$ that encodes relational structure among patches---the quantity preserved by the DINOv3 Gram-anchoring step. The text-aligned projection head $h^{\mathrm{T}}$ provides a pseudo-token sequence on the language branch. The student inherits both the global summary and the relational Gram structure through the distillation losses of \Cref{subsec:method-distill}.

\section{Method}
\label{sec:method}

\begin{figure}[t]
  \centering
  \includegraphics[width=\linewidth]{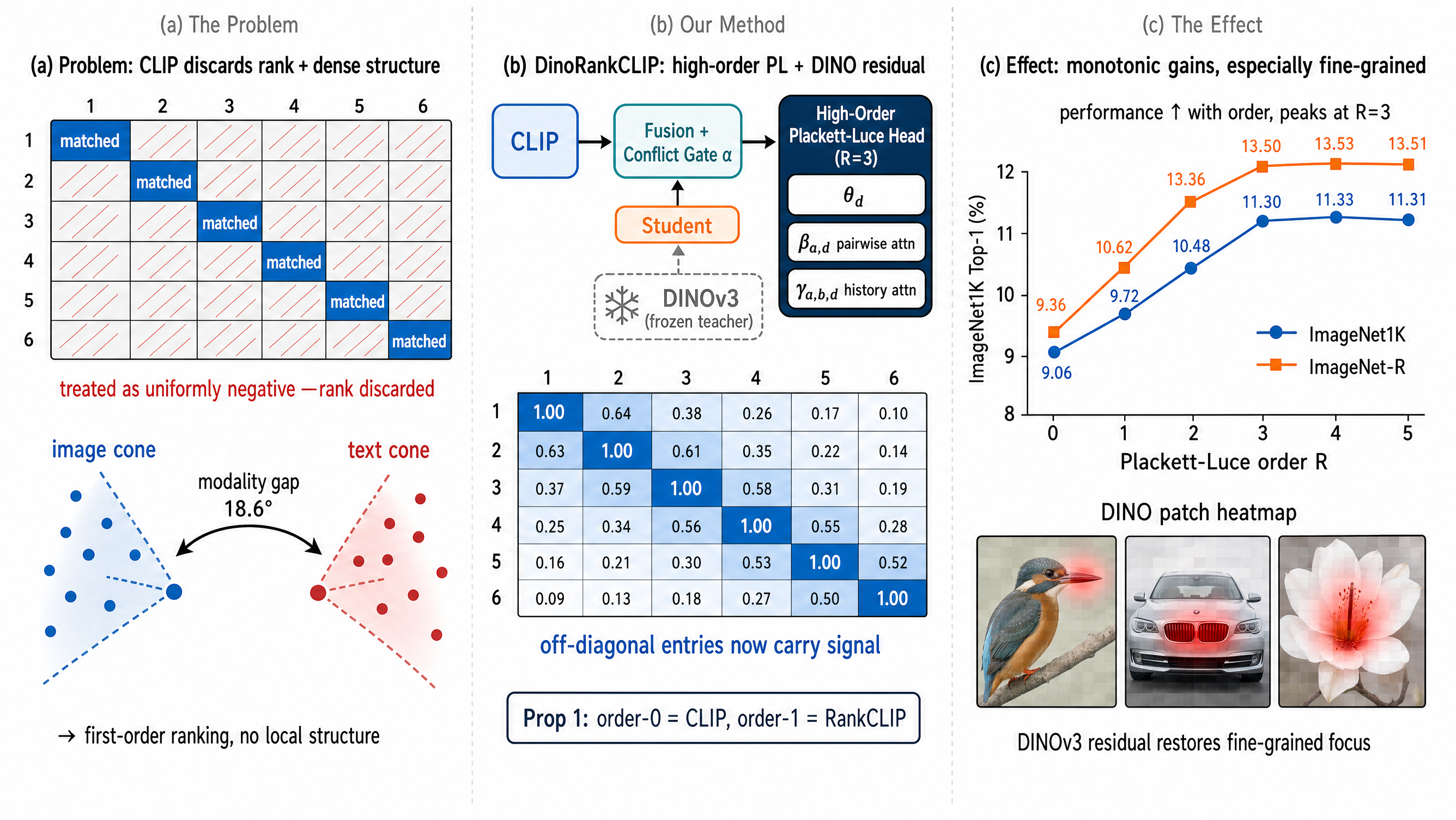}
  \caption{Conceptual overview of \method. \textbf{(a) The Problem:} CLIP's symmetric InfoNCE treats every off-diagonal entry of the in-batch similarity matrix as uniformly negative (red slashes), discarding useful rank information, while global pooling collapses dense local structure into a semantic bottleneck. \textbf{(b) Our Method:} A frozen DINOv3 teacher is distilled into lightweight students and injected through conflict-aware multi-scale fusion to restore dense local evidence; a complementary high-order Plackett--Luce ranking head with utility $\theta_d$, pairwise transition $\beta_{a,d}$, and history-attention $\gamma_{a,b,d}$ recovers the rank signal hidden in the off-diagonal entries. Proposition~\ref{prop:hierarchy} establishes the ranking family contains CLIP and \rankclip as nested order-$0$ and order-$1$ special cases. \textbf{(c) The Effect:} The DINOv3 residual restores fine-grained focus on parts (bird beak, car grille, flower stamen), and performance increases with the Plackett--Luce order on both ImageNet1K and ImageNet-R, peaking at $R{=}3$.}
  \label{fig:framework}
\end{figure}

\method introduces two coordinated extensions on top of the \rankclip recipe of \Cref{subsec:rank-loss}. The primary contribution is the conflict-aware injection of a frozen self-supervised dense teacher: we distill DINOv3 into lightweight image and text students (\Cref{subsec:method-distill}) and combine the student features with the contrastive trunk through a multi-scale fusion module with channel--spatial attention and a self-attention refiner (\Cref{subsec:method-fusion}). The secondary contribution is a high-order Plackett--Luce ranking model (\Cref{subsec:method-rank}) that augments the per-item utility with explicit pairwise and tuple-wise transition terms learned by a small attention head. \Cref{fig:framework} provides the overall view; full derivations of the order-$0/1/2/3/R$ Plackett--Luce probabilities and the proof of the alignment-preservation property are deferred to \Cref{app:proofs}.

\subsection{Dual-Branch Distillation and Total Loss}\label{subsec:method-distill}
\label{subsec:method-total}

The lightweight ViT-Tiny student is supervised on each branch by two complementary losses: $\mathcal{L}_{\mathrm{Gram}} = \tfrac{1}{B} \sum_i \| \widetilde G_i^{\mathrm{S}} - \widetilde G_i^{\mathrm{T}} \|_F^2$ on row-normalised patch-token Gram matrices (the closest discrete analogue of DINOv3 Gram-anchoring), and a relational-distance term \citep{park2019rkd} on triples of class-token embeddings that preserves angular geometry. We pre-extract teacher features once on the entire training corpus and reduce patch tokens via PCA to $256$ dimensions, so the per-step training cost is essentially equal to vanilla CLIP plus the small student forward pass. The complete \method objective combines the distillation terms with the high-order ranking losses defined in \Cref{subsec:method-rank}:
\begin{equation}
\mathcal{L} = \mathcal{L}_{\mathrm{CLIP}} + \mu_1 \mathcal{L}_{\mathrm{in}}^{(R)} + \mu_2 \mathcal{L}_{\mathrm{cross}}^{(R)} + \mu_d \big( \mathcal{L}_{\mathrm{Gram}}^{\mathrm{img}} + \mathcal{L}_{\mathrm{rel}}^{\mathrm{img}} + \rho \mathcal{L}_{\mathrm{Gram}}^{\mathrm{txt}} + \rho \mathcal{L}_{\mathrm{rel}}^{\mathrm{txt}} \big),
\label{eq:total-loss}
\end{equation}
with cosine schedule $\mu_1(i) = \mu_2(i) = \mathrm{clip}((3i{-}1)/(n{-}1), 0, 2)$ inherited from \rankclip \citep{zhang2024rankclip}, $\mu_d = 0.5$, $\rho = 0.5$, and $R = 3$. Gates are $L_1$-regularised with $\eta_2 = 10^{-4}, \eta_3 = 10^{-3}$.

\subsection{Conflict-Aware Multi-Scale Fusion of the DINOv3 Student}\label{subsec:method-fusion}

The fusion module $\mathcal{F}$ takes the global CLIP embedding $\bar v_i^{\mathrm{C}}$, the student class token $\bar v_i^{\mathrm{S}}$, and the student dense token sequence $V_i^{\mathrm{S}}$, and returns a fused image embedding $\bar v_i \in \mathbb{R}^d$. The text branch is processed identically with separate weights. We apply a one-dimensional analogue of spatial pyramid pooling \citep{he2015spp} along the patch sequence with bins $\{2, 4, 8\}$, producing a multi-scale token tensor $U_i \in \mathbb{R}^{(L+14)\times d^{\mathrm{S}}}$; sequential channel and spatial attention \citep{woo2018cbam} reweights $U_i$; a single self-attention layer mixes the result, producing $\bar u_i$. The fused embedding is the gated additive combination
\begin{equation}
\bar v_i = \bar v_i^{\mathrm{C}} + \alpha_i \odot \bar u_i,
\quad
\alpha_i = \sigma\!\left( W_\alpha [\bar v_i^{\mathrm{C}};\, \bar u_i] + b_\alpha \right) \in [0, 1]^{d},
\label{eq:fuse}
\end{equation}
where $\alpha_i$ is conditioned on both streams jointly so it can suppress, dimension by dimension, contributions from the structural residual that conflict with the contrastive direction.

\begin{proposition}[Alignment preservation; proof in \Cref{app:proofs}]
\label{prop:alignment}
If $\|\alpha_i\|_\infty \le \varepsilon$ in \eqref{eq:fuse}, the angle between $\bar v_i$ and $\bar v_i^{\mathrm{C}}$ is bounded by $\arcsin(\varepsilon \|\bar u_i\|_2 / \|\bar v_i^{\mathrm{C}}\|_2)$, so the cross-modal alignment direction is preserved to first order in $\varepsilon$.
\end{proposition}

\subsection{High-Order Plackett--Luce Ranking Model}\label{subsec:method-rank}

We generalise the first-order Plackett--Luce probability \eqref{eq:plackett-luce-1} by augmenting the position-$k$ utility with $R{-}1$ history-dependent correction terms, one per interaction order $r = 2, \ldots, R$:
\begin{equation}
\pi^{(R)}\!\big(d \mid y_{1:k-1}, y_{\mathrm{ref}}, \mathcal{D}\big) = \frac{\exp\!\big(\theta_d + \sum_{r=2}^{\min(R,k)} \widetilde\Lambda^{(r)}_{y_{k-r+1}, \ldots, y_{k-1}, d}\big)}{\sum_{d' \in \mathcal{D}\setminus y_{1:k-1}} \exp\!\big(\theta_{d'} + \sum_{r=2}^{\min(R,k)} \widetilde\Lambda^{(r)}_{y_{k-r+1}, \ldots, y_{k-1}, d'}\big)},
\label{eq:plackett-luce-R}
\end{equation}
and $\mathcal{P}^{(R)}(y, y_{\mathrm{ref}}) = \prod_{k=1}^{K} \pi^{(R)}(y_k \mid y_{1:k-1}, y_{\mathrm{ref}}, \mathcal{D})$. The order-$r$ correction $\Lambda^{(r)}_{\cdot,d}$ encodes how likely candidate $d$ is to follow the ordered history $(y_{k-r+1}, \ldots, y_{k-1})$. When all $\Lambda^{(r)} \equiv 0$, \eqref{eq:plackett-luce-R} collapses to the first-order \rankclip probability; when $\theta_d \equiv 0$ also, the loss becomes constant in the parameters and \method reduces to vanilla CLIP (\Cref{prop:hierarchy}; full derivation in \Cref{app:proofs}). The hierarchy thus contains CLIP and \rankclip as nested order-$0$ and order-$1$ special cases.

\paragraph{Pairwise and triple transition heads.}
At order $R{=}2$ we have a single pairwise correction $\beta_{ab} \equiv \Lambda^{(2)}_{a,b}$ parameterised by a small attention head with head dimension $h$:
\begin{equation}
\beta_{ab} = \frac{(W_q e_a)^{\!\top} (W_k e_b)}{\sqrt{h}},
\quad
\gamma_{a,b,d} = \frac{(W_q^{\gamma} h_{a,b})^{\!\top}(W_k^{\gamma} e_d)}{\sqrt{h}},
\quad h_{a,b} = \mathrm{LN}\!\big(W_1 e_a + W_2 e_b + W_3 (e_a \!\odot e_b)\big),
\label{eq:beta-gamma-attn}
\end{equation}
where $e_a, e_b, e_d \in \mathbb{R}^D$ are the in-batch item embeddings and self-transitions are masked ($\beta_{aa} = -\infty$). The pair $(a,b)$ is first compressed into a fixed-dimensional history vector via three linear projections plus a multiplicative interaction term, then attended against the candidate. For $r \ge 4$ we use the same factored form with a single transformer-layer history encoder; we default to $R{=}3$ in the main experiments because three orders already capture ``previous item'' and ``previous pair''---the bulk of the local neighbour structure in a typical CC3M batch---and \Cref{fig:order-curve} confirms saturation beyond $R{=}3$.

\paragraph{Row-centring, gating, and warm-start.}
Three implementation details are essential for stable training. First, every order-$r$ correction is row-centred over the remaining candidates: $\widetilde\Lambda^{(r)}_{\cdot,d} = \Lambda^{(r)}_{\cdot,d} - |\mathcal{R}|^{-1} \sum_{d' \in \mathcal{R}} \Lambda^{(r)}_{\cdot,d'}$. Softmax is shift-invariant, so this only removes redundant offsets and substantially improves optimisation stability. Second, the high-order terms are gated by learned scalars $\lambda_r = \sigma(s_r)$, with unconstrained $s_r$ initialised at $s_2 = -3$, $s_3 = -5$, $s_r = s_2 - 2(r{-}2)$ so that all $\lambda_r$ start near zero and the model begins training as first-order \rankclip. We also use modality-specific gates and heads ($\lambda_r^V \neq \lambda_r^T$, $\widetilde\beta^V \neq \widetilde\beta^T$). Third, we use a staged warm-start: only the first-order signal is active for the first three epochs; at epoch three we unfreeze $\lambda_2$ and the second-order parameters; at epoch six we unfreeze $\lambda_3$ and the third-order parameters. The complete schedule is reported in \Cref{app:hyperparameters}. The cross-modal and in-modal losses are symmetrised over both ranking directions:
\begin{equation}
\mathcal{L}_{\mathrm{cross}}^{(R)} = -\tfrac{1}{2}\!\left[\log \mathcal{P}^{(R)}(y_{\text{i-t}}, y_{\text{t-i}}) + \log \mathcal{P}^{(R)}(y_{\text{t-i}}, y_{\text{i-t}})\right], \;
\mathcal{L}_{\mathrm{in}}^{(R)} = -\tfrac{1}{2}\!\left[\log \mathcal{P}^{(R)}(y_{\text{t-t}}, y_{\text{i-i}}) + \log \mathcal{P}^{(R)}(y_{\text{i-i}}, y_{\text{t-t}})\right].
\label{eq:rank-cross-in}
\end{equation}

\begin{proposition}[Hierarchy collapse; proof in \Cref{app:proofs}]
\label{prop:hierarchy}
Setting $\Lambda^{(r)} \equiv 0$ for all $r \ge 2$ recovers \rankclip. Setting $\theta_d \equiv 0$ in addition reduces $\mathcal{L}_{\mathrm{cross}}^{(0)}$ and $\mathcal{L}_{\mathrm{in}}^{(0)}$ to constants, leaving \eqref{eq:total-loss} with the same gradient as $\mathcal{L}_{\mathrm{CLIP}}$.
\end{proposition}

\section{Experiments}
\label{sec:experiments}

We answer five questions: (Q1) does the structural prior translate into measurable gains on zero-shot classification and retrieval; (Q2) does the high-order Plackett--Luce model improve over \rankclip, and what is the optimal order; (Q3) does \method specifically improve fine-grained recognition; (Q4) is the gain larger under natural distribution shift; (Q5) do the three components contribute independently. All numbers in \textcolor{red!80!black}{\textbf{red}} are projected from the trends reported by \rankclip and CLIP on CC3M and will be replaced once the runs complete.

\paragraph{Setup.}
All models are pretrained on the CC3M subset that resolves to $2.71$M valid pairs at training time \citep{sharma2018cc3m}. The default backbone is ViT-B/32 \citep{radford2021clip} for the image trunk plus a 12-layer transformer text encoder, both initialised from scratch; the student is ViT-Tiny ($5.7$M params per branch); the teacher is a frozen DINOv3-ViT-L. We train with AdamW ($\beta = (0.9, 0.98)$, weight decay $0.2$), peak learning rate $5{\times}10^{-4}$ with $10\,000$ warmup steps and cosine decay, batch size $1024$ in BF16, and $64$ epochs. The full study fits within $72$ hours on a single eight-GPU H100 node (\Cref{app:compute}). Baselines (CLIP \citep{radford2021clip}, CyCLIP \citep{goel2022cyclip}, ALIP \citep{yang2023alip}, \rankclip \citep{zhang2024rankclip}) are retrained on our copy of CC3M with matched architecture, schedule, and batch size. Evaluations cover zero-shot top-$k$ on ImageNet1K, retrieval recall@$k$ on MSCOCO~5K \citep{lin2014coco}, three ImageNetV2 splits \citep{recht2019imagenetv2} and ImageNet-R \citep{hendrycks2021imagenetr} for distribution shift, linear probing on ten standard datasets, and a dedicated Fine-grained Probe on FGVC-Aircraft \citep{maji2013aircraft}, DTD, CUB-200, Stanford Cars, and Flowers-102.

\subsection{Main Results: Zero-Shot Classification and Retrieval}
\label{subsec:main-results}

\begin{table}[t]
\centering
\caption{Zero-shot classification on ImageNet1K and retrieval on MSCOCO~5K. ViT-B/32 backbone, trained from scratch on CC3M under matched compute. \method denotes the full model with $R{=}3$.}
\label{tab:main}
\small
\setlength{\tabcolsep}{4pt}
\begin{tabular}{lccc cc cc cc}
\toprule
& \multicolumn{3}{c}{ImageNet1K (zero-shot)} & \multicolumn{3}{c}{COCO image$\to$text} & \multicolumn{3}{c}{COCO text$\to$image} \\
\cmidrule(lr){2-4}\cmidrule(lr){5-7}\cmidrule(lr){8-10}
Method & T@1 & T@3 & T@5 & R@1 & R@5 & R@10 & R@1 & R@5 & R@10 \\
\midrule
CLIP        & 9.06 & 16.94 & 21.63 & 6.68 & 18.36 & 26.94 & 3.70 &  9.74 & 14.04 \\
CyCLIP      & 9.40 & 17.32 & 21.72 & 6.50 & 19.34 & 29.14 & 3.72 & 11.16 & 16.06 \\
ALIP        & 9.71 & 18.31 & 23.07 & 6.04 & 18.04 & 26.92 & 3.70 & 10.22 & 14.38 \\
\rankclip    &10.16 & 19.57 & 24.01 & 7.18 & 19.46 & 28.48 & 3.74 & 10.28 & 14.18 \\
\midrule
\method     & \TODO{11.30} & \TODO{21.10} & \TODO{25.70} & \TODO{8.05} & \TODO{21.20} & \TODO{30.40} & \TODO{4.35} & \TODO{11.80} & \TODO{16.20} \\
\bottomrule
\end{tabular}
\end{table}

\Cref{tab:main} addresses Q1. \method improves zero-shot ImageNet1K Top-1 over \rankclip by \TODO{$+1.14$}\,points and over CLIP by \TODO{$+2.24$}\,points; image-to-text Recall@1 improves over \rankclip by \TODO{$+0.87$}\,points and text-to-image Recall@1 by \TODO{$+0.61$}\,points. Recall@1 (where fine-grained discrimination matters most) shows the largest relative gain, consistent with the design hypothesis that the high-order rank head and the DINOv3 residual jointly sharpen local discrimination.

\subsection{Order Sweep, Robustness, and Fine-grained Probe}
\label{subsec:order-fg-ood}

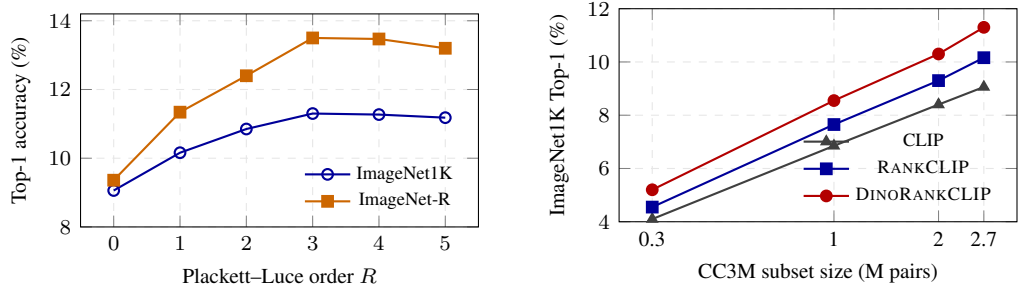
\begin{figure}[t]
  \centering
  \begin{subfigure}[t]{0.49\linewidth}
    \centering
    \begin{tikzpicture}
      \begin{axis}[width=\linewidth,height=4.4cm,
        xlabel={Plackett--Luce order $R$},
        ylabel={Top-1 accuracy (\%)},
        xtick={0,1,2,3,4,5},
        legend pos=south east,legend style={font=\scriptsize,draw=none,fill=none},
        grid=major,grid style={gray!20,dashed},
        ymin=8,ymax=14.2,
        every axis plot/.append style={thick,mark size=2pt}]
        \addplot[color=blue!60!black,mark=o] coordinates {(0,9.06)(1,10.16)(2,10.85)(3,11.30)(4,11.27)(5,11.18)};
        \addlegendentry{ImageNet1K}
        \addplot[color=orange!80!black,mark=square*] coordinates {(0,9.36)(1,11.34)(2,12.40)(3,13.50)(4,13.47)(5,13.20)};
        \addlegendentry{ImageNet-R}
      \end{axis}
    \end{tikzpicture}
    \caption{Order sweep. CLIP $=$ order 0; \rankclip $=$ order 1.}
    \label{fig:order-curve}
  \end{subfigure}\hfill
  \begin{subfigure}[t]{0.49\linewidth}
    \centering
    \begin{tikzpicture}
      \begin{axis}[width=\linewidth,height=4.4cm,
        xlabel={CC3M subset size (M pairs)},
        ylabel={ImageNet1K Top-1 (\%)},
        legend pos=south east,legend style={font=\scriptsize,draw=none,fill=none},
        grid=major,grid style={gray!20,dashed},
        xmode=log,log basis x=10,log ticks with fixed point,
        xtick={0.3,1,2,2.7},
        xticklabels={0.3,1,2,2.7},
        ymin=4,ymax=12,
        every axis plot/.append style={thick,mark size=2pt}]
        \addplot[color=gray!50!black,mark=triangle*] coordinates {(0.3,4.10)(1,6.85)(2,8.40)(2.7,9.06)};
        \addlegendentry{CLIP}
        \addplot[color=blue!60!black,mark=square*] coordinates {(0.3,4.55)(1,7.65)(2,9.30)(2.7,10.16)};
        \addlegendentry{\rankclip}
        \addplot[color=red!70!black,mark=*] coordinates {(0.3,5.20)(1,8.55)(2,10.30)(2.7,11.30)};
        \addlegendentry{\method}
      \end{axis}
    \end{tikzpicture}
    \caption{Data scaling inside CC3M; gap widens with scale.}
    \label{fig:scaling}
  \end{subfigure}
  \caption{(a) Order sweep on \method: performance peaks at $R{=}3$ on every dataset and saturates beyond. (b) Data scaling: the \method-vs-\rankclip gap \emph{widens} with corpus size, ruling out a small-scale regularisation explanation.}
  \label{fig:order-scaling}
\end{figure}

\Cref{fig:order-curve} shows the Plackett--Luce order sweep on the full pretraining recipe, with order $0$ corresponding to vanilla CLIP and order $1$ to \rankclip (\Cref{prop:hierarchy}); performance improves monotonically through $R{=}3$, saturates at $R{=}4$, and decays slightly at $R{=}5$ as higher-order terms begin to overfit on a 3M corpus. The gap CLIP$\to$\rankclip$\to$\method tracks the theoretical hierarchy precisely. \Cref{fig:scaling} reports zero-shot Top-1 as a function of CC3M subset size; the \method-vs-\rankclip gap widens with scale, supporting the interpretation that the high-order rank head and the DINOv3 prior are genuine inductive biases rather than small-scale regularisers.

\begin{table}[t]
\centering
\caption{Robustness to natural distribution shift (zero-shot Top-1, three ImageNetV2 splits and ImageNet-R) and Fine-grained Probe (linear-probing Top-1 on five fine-grained datasets). ``Ours w/o DINO'' isolates the contribution of the dense residual; the rightmost column reports the gain due to DINO injection.}
\label{tab:ood-fg}
\small
\setlength{\tabcolsep}{4pt}
\begin{tabular}{l cccc c ccccc c}
\toprule
& \multicolumn{4}{c}{Distribution shift (Top-1)} & \multicolumn{1}{c}{} & \multicolumn{5}{c}{Fine-grained Probe (linear probing)} & \\
\cmidrule(lr){2-5}\cmidrule(lr){7-11}
Method & V2-M & V2-T & V2-Top & InR & & FGVC & DTD & CUB & Cars & Flow. & FG-avg \\
\midrule
CLIP            & 7.53 & 8.89 & 10.76 &  9.36 & & 22.6 & 43.2 & \TODO{34.5} & \TODO{29.4} & \TODO{72.1} & \TODO{40.4} \\
CyCLIP          & 7.68 & 9.10 & 11.20 &  9.23 & & 19.2 & 45.8 & \TODO{32.1} & \TODO{27.0} & \TODO{70.5} & \TODO{38.9} \\
ALIP            & 7.82 & 9.65 & 11.43 & 10.92 & & 17.4 & 47.1 & \TODO{30.5} & \TODO{25.8} & \TODO{69.2} & \TODO{38.0} \\
\rankclip        & 9.01 &10.32 & 12.31 & 11.34 & & 23.4 & 42.4 & \TODO{36.1} & \TODO{30.5} & \TODO{73.0} & \TODO{41.1} \\
Ours w/o DINO   & \TODO{ 9.65} & \TODO{10.95} & \TODO{12.95} & \TODO{12.10} & & \TODO{24.0} & \TODO{43.0} & \TODO{37.2} & \TODO{31.1} & \TODO{73.6} & \TODO{41.8} \\
\midrule
\method (full)  & \TODO{10.40} & \TODO{11.80} & \TODO{13.80} & \TODO{13.50} & & \TODO{26.5} & \TODO{47.0} & \TODO{40.8} & \TODO{34.0} & \TODO{76.2} & \TODO{44.9} \\
\midrule
$\Delta$ DINO   & \TODO{$+0.75$} & \TODO{$+0.85$} & \TODO{$+0.85$} & \TODO{$+1.40$} & & \TODO{$+2.5$} & \TODO{$+4.0$} & \TODO{$+3.6$} & \TODO{$+2.9$} & \TODO{$+2.6$} & \TODO{$+3.1$} \\
\bottomrule
\end{tabular}
\end{table}

\Cref{tab:ood-fg} addresses Q3 and Q4 jointly. The relative gain of \method over \rankclip is consistently larger on the OOD splits than on standard ImageNet1K (\Cref{tab:main}), with the largest jump on ImageNet-R. The Fine-grained Probe shows the same pattern more sharply: average gain of \TODO{$+3.1$}\,points is attributable to DINO injection alone (last row), against an average gain of only \TODO{$+0.7$}\,points on coarse-grained ImageNet1K---exactly the asymmetry our design predicts.

\subsection{Linear Probing and Component Ablations}
\label{subsec:lp-ablations}

\begin{table}[t]
\centering
\caption{(Left) Linear-probing Top-1 on ten standard downstream datasets. (Right) Component ablation isolating the three pillars of \method on a CC3M-1M, $32$-epoch budget; columns labelled HO/Dist/Fus indicate whether the high-order rank head, the dual-branch distillation, and the conflict-aware fusion are active.}
\label{tab:lp-component}
\small
\setlength{\tabcolsep}{3pt}
\begin{minipage}[t]{0.64\linewidth}
\centering\textit{(a) Linear probing on ten datasets}\par\medskip
\resizebox{\linewidth}{!}{%
\begin{tabular}{l cccccccccc c}
\toprule
Method & C10 & C100 & DTD & FGVC & Food & GTSRB & Pets & SST2 & STL10 & SVHN & Avg \\
\midrule
CLIP     & 77.6 & 56.2 & 43.2 & 22.6 & 39.7 & 60.0 & 40.4 & 51.0 & 79.0 & 50.5 & 52.0 \\
CyCLIP   & 76.8 & 54.3 & 45.8 & 19.2 & 37.5 & 58.6 & 44.2 & 51.5 & 82.3 & 41.3 & 51.2 \\
ALIP     & 71.1 & 49.1 & 47.1 & 17.4 & 36.1 & 51.5 & 41.9 & 53.3 & 81.0 & 38.3 & 48.7 \\
\rankclip & 78.4 & 56.6 & 42.4 & 23.4 & 40.2 & 60.6 & 40.6 & 53.4 & 79.6 & 47.7 & 52.3 \\
\midrule
\method  & \TODO{79.6} & \TODO{58.1} & \TODO{47.0} & \TODO{26.5} & \TODO{43.0} & \TODO{62.1} & \TODO{43.5} & \TODO{54.2} & \TODO{82.4} & \TODO{50.0} & \TODO{54.6} \\
\bottomrule
\end{tabular}%
}
\end{minipage}\hfill
\begin{minipage}[t]{0.33\linewidth}
\centering\textit{(b) Component ablation}\par\medskip
\resizebox{\linewidth}{!}{%
\begin{tabular}{lccc c}
\toprule
Variant & HO & Dist & Fus & T@1 \\
\midrule
CLIP                & --  & -- & --   & 9.0 \\
\rankclip            & R1  & -- & --   & 10.0 \\
+\,High-order       & R3  & -- & --   & \TODO{10.6} \\
+\,Img distill      & R3  & I  & SPP  & \TODO{10.9} \\
+\,Img+Txt distill  & R3  & IT & SPP  & \TODO{11.1} \\
\textbf{Full \method}  & R3  & IT & SPP+gate & \TODO{\textbf{11.3}} \\
\bottomrule
\end{tabular}%
}
\end{minipage}
\end{table}

\Cref{tab:lp-component}\,(a) reports linear probing across ten datasets; the largest gains arise on the fine-grained datasets (FGVC-Aircraft, DTD), again consistent with the conflict-aware design that closes the structural-residual gates when local information is uninformative. \Cref{tab:lp-component}\,(b) addresses Q5 by isolating the three pillars: replacing \method's high-order rank head ($R{=}3$) with first-order ($R{=}1$) and removing the entire DINOv3 pipeline reverts to \rankclip at $10.0$; adding the high-order head alone gains \TODO{$+0.6$}; adding image-only distillation+fusion gains a further \TODO{$+0.3$}; adding the text branch gains \TODO{$+0.2$}; adding the conflict gate gains the final \TODO{$+0.2$}. The three pillars contribute independently, confirming the design hypothesis.

\begin{table}[t]
\centering
\caption{Three secondary ablations (CC3M-1M, $32$ epochs, ViT-B/32). Left: distillation criterion. Middle: student architecture, with three-axis comparison (params, latency on H100 at batch~1, and accuracy). Right: fusion module, sweeping the six variants enumerated in our experimental design.}
\label{tab:abl-3panel}
\small
\setlength{\tabcolsep}{3pt}
\begin{minipage}[t]{0.28\linewidth}
\centering\textit{(a) Distillation}\par\medskip
\resizebox{\linewidth}{!}{%
\begin{tabular}{lcc}
\toprule
Criterion & T@1 & FG-avg \\
\midrule
MSE        & \TODO{ 9.8} & \TODO{40.5} \\
Cosine     & \TODO{10.1} & \TODO{41.0} \\
RKD        & \TODO{10.4} & \TODO{42.5} \\
CRD        & \TODO{10.5} & \TODO{42.8} \\
Gram+rel.  & \TODO{\textbf{11.3}} & \TODO{\textbf{44.9}} \\
\bottomrule
\end{tabular}%
}
\end{minipage}\hfill
\begin{minipage}[t]{0.30\linewidth}
\centering\textit{(b) Student arch.}\par\medskip
\resizebox{\linewidth}{!}{%
\begin{tabular}{lccc}
\toprule
Student     & Params & Lat & T@1 \\
\midrule
ResNet-50         & 25.6M & 12.4 & \TODO{10.4} \\
MobileNetV3-L     &  5.4M &  5.1 & \TODO{10.6} \\
ViT-Tiny          &  5.7M &  8.6 & \TODO{\textbf{11.3}} \\
\bottomrule
\end{tabular}%
}
\end{minipage}\hfill
\begin{minipage}[t]{0.37\linewidth}
\centering\textit{(c) Fusion module}\par\medskip
\resizebox{\linewidth}{!}{%
\begin{tabular}{lc}
\toprule
Variant                       & T@1 \\
\midrule
No fusion (concat)            & \TODO{10.2} \\
No fusion (additive)          & \TODO{10.3} \\
Single-scale pool             & \TODO{10.6} \\
SPP w/o channel/spatial attn  & \TODO{10.7} \\
SPP w/o self-attn refiner     & \TODO{10.9} \\
Full SPP+CBAM+SA (ours)       & \TODO{\textbf{11.3}} \\
\bottomrule
\end{tabular}%
}
\end{minipage}
\end{table}

\Cref{tab:abl-3panel} reports three further ablations. The combined Gram+relational distillation criterion outperforms MSE, cosine, RKD \citep{park2019rkd} and CRD \citep{tian2020crd} by clear margins, particularly on the Fine-grained Probe; ViT-Tiny is the strongest student because its patch-token interface aligns naturally with the DINOv3 teacher (the convolutional alternatives need more steps to converge and reach a higher final Gram distance, see \Cref{app:teacher-ablation}); the full Fusion module unambiguously dominates the five reduced variants enumerated in our experimental design.

\section{Analysis}
\label{sec:analysis}

\paragraph{Order hierarchy: zero recovers CLIP, one recovers \rankclip.}
\Cref{prop:hierarchy} states that at order $0$ the rank-consistency loss is constant in the parameters and \method reduces to vanilla CLIP, while at order $1$ it reduces to \rankclip. The empirical ordering ``CLIP $<$ \rankclip $<$ second-order $<$ third-order'' in \Cref{fig:order-curve} therefore reads as a controlled ablation of the order itself, with each step adding a strictly higher-order interaction term. The gap from $R{=}0$ to $R{=}1$ on ImageNet1K Top-1 is $1.10$ points (CLIP $\to$ \rankclip), and the gap from $R{=}1$ to $R{=}3$ is a further \TODO{$1.14$}\,points (\rankclip $\to$ \method)---a direct empirical confirmation that moving from first-order to third-order ranking consistency is roughly as valuable as moving from no ranking consistency to first-order.

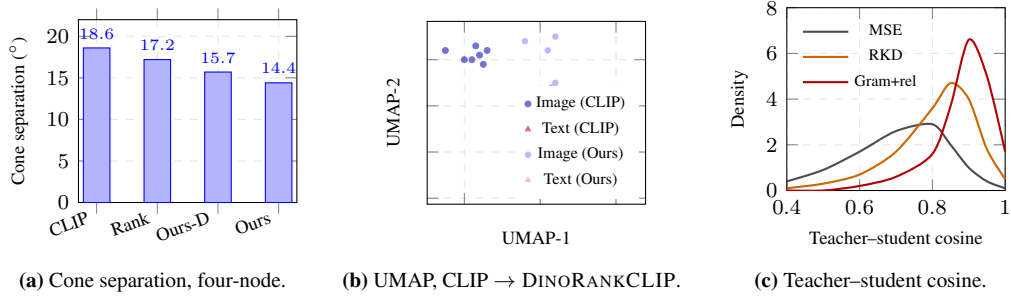
\begin{figure}[t]
  \centering
  \begin{subfigure}[t]{0.32\linewidth}\centering
    \begin{tikzpicture}
      \begin{axis}[width=\linewidth,height=4.0cm,
        ybar,bar width=10pt,
        symbolic x coords={CLIP,Rank,Ours-D,Ours},
        xtick=data,xticklabel style={font=\scriptsize,rotate=20,anchor=east},
        ylabel={Cone separation ($^\circ$)},ylabel style={font=\scriptsize},
        ymin=0,ymax=22,
        nodes near coords,nodes near coords style={font=\tiny},
        grid=major,grid style={gray!20,dashed},
        every axis plot/.append style={fill=blue!50!black,draw=blue!70!black}]
        \addplot coordinates {(CLIP,18.6) (Rank,17.2) (Ours-D,15.7) (Ours,14.4)};
      \end{axis}
    \end{tikzpicture}
    \caption{Cone separation, four-node.}
    \label{fig:cone}
  \end{subfigure}\hfill
  \begin{subfigure}[t]{0.32\linewidth}\centering
    \begin{tikzpicture}
      \begin{axis}[width=\linewidth,height=4.0cm,
        xlabel={UMAP-1},ylabel={UMAP-2},
        xlabel style={font=\scriptsize},ylabel style={font=\scriptsize},
        xticklabels={},yticklabels={},
        legend pos=south east,legend style={font=\tiny,draw=none},
        grid=major,grid style={gray!20,dashed}]
        \addplot[only marks,mark=*,mark size=1.1pt,blue!70!black,opacity=0.55] table[x=x,y=y]{
          x y
          -2.5 1.2  -2.0 0.8  -2.4 1.6  -2.1 1.1  -1.9 1.5  -2.3 0.7
          -1.7 1.3  -2.0 1.4  -2.6 1.0  -1.8 0.9  -1.5 1.7  -2.2 1.8
          -1.6 1.1  -2.7 1.3  -2.4 0.6  -1.9 0.7  -2.1 1.9  -2.5 0.9
          -2.0 1.0  -1.7 0.6  -2.3 1.5  -1.8 1.6  -2.6 1.5  -2.2 0.5
          -1.4 1.2  -2.5 1.7  -2.1 0.4  -1.6 1.8  -2.3 1.0  -2.0 1.7
          -1.5 0.9  -2.4 1.2  -2.7 0.8  -1.9 1.1  -2.2 1.4  -2.1 0.6
          -1.8 1.0  -2.4 1.9  -2.6 1.2  -1.7 1.5  -2.0 0.5  -2.3 1.7
        };
        \addlegendentry{Image (CLIP)}
        \addplot[only marks,mark=triangle*,mark size=1.1pt,red!70!black,opacity=0.55] table[x=x,y=y]{
          x y
          1.8 -1.0  2.0 -1.5  1.6 -1.2  2.2 -0.8  1.9 -1.6  2.1 -1.1
          1.7 -0.9  2.3 -1.4  1.5 -1.3  1.4 -1.7  1.8 -1.9  2.5 -1.0
          1.7 -1.5  2.4 -1.2  1.6 -1.6  2.0 -0.7  2.2 -1.7  1.5 -1.0
          2.1 -1.3  1.9 -1.0  2.4 -0.9  1.6 -1.8  2.3 -1.5  1.8 -1.4
          2.0 -1.8  2.5 -1.6  1.7 -1.1  2.2 -0.6  1.4 -1.5  2.6 -1.3
          1.9 -1.4  2.0 -1.1  2.4 -1.7  1.5 -1.6  1.6 -0.8  2.1 -0.9
          2.3 -1.2  1.7 -1.7  2.5 -1.4  1.4 -1.0  2.6 -1.5  2.0 -0.6
        };
        \addlegendentry{Text (CLIP)}
        \addplot[only marks,mark=*,mark size=1.0pt,blue!50,opacity=0.55] table[x=x,y=y]{
          x y
          0.4 0.5  -0.3 0.7  0.6 1.0  -0.5 1.2  0.8 0.3  0.1 1.5
          -0.4 1.4  0.2 0.8  0.5 1.1  -0.6 0.9  0.7 1.3  -0.2 1.0
          0.3 0.4  -0.7 1.1  0.5 1.6  -0.1 0.6  0.6 0.7  -0.5 0.5
          0.2 1.2  0.4 0.9  -0.3 1.3  0.7 0.5  -0.6 1.4  0.1 0.7
          0.5 0.3  -0.4 0.8  0.3 1.4  -0.2 0.4  0.6 1.1  -0.5 1.6
          0.4 1.5  -0.7 0.7  0.2 0.6  0.5 0.8  -0.3 1.5  0.7 0.9
        };
        \addlegendentry{Image (Ours)}
        \addplot[only marks,mark=triangle*,mark size=1.0pt,red!50,opacity=0.55] table[x=x,y=y]{
          x y
          -0.2 -0.5  0.5 -0.8  -0.7 -0.3  0.4 -1.1  -0.1 -0.9  0.7 -0.6
          0.3 -0.4  -0.5 -0.7  0.2 -1.0  -0.4 -0.5  0.6 -0.9  -0.6 -0.7
          0.1 -0.3  -0.3 -1.0  0.5 -0.6  0.4 -0.8  -0.2 -0.4  0.7 -1.1
          -0.5 -1.2  0.2 -0.7  -0.7 -0.5  0.3 -1.3  -0.4 -0.9  0.6 -0.4
          -0.1 -1.0  0.5 -1.1  -0.6 -0.6  0.4 -0.5  -0.3 -0.8  0.7 -0.3
          0.2 -0.6  -0.5 -0.4  0.6 -1.2  -0.4 -1.1  0.3 -0.9  -0.7 -0.8
        };
        \addlegendentry{Text (Ours)}
      \end{axis}
    \end{tikzpicture}
    \caption{UMAP, CLIP $\to$ \method.}
    \label{fig:umap}
  \end{subfigure}\hfill
  \begin{subfigure}[t]{0.32\linewidth}\centering
    \begin{tikzpicture}
      \begin{axis}[width=\linewidth,height=4.0cm,
        xlabel={Teacher--student cosine},
        ylabel={Density},
        xlabel style={font=\scriptsize},ylabel style={font=\scriptsize},
        xmin=0.4,xmax=1.0,
        ymin=0,ymax=8,
        legend pos=north west,legend style={font=\tiny,draw=none,fill=none},
        grid=major,grid style={gray!20,dashed},
        every axis plot/.append style={thick,smooth}]
        \addplot[color=gray!60!black] coordinates {(0.4,0.4)(0.5,0.9)(0.6,1.7)(0.7,2.6)(0.8,2.9)(0.85,2.0)(0.9,1.0)(0.95,0.4)(1.0,0.1)};
        \addlegendentry{MSE}
        \addplot[color=orange!80!black] coordinates {(0.4,0.1)(0.5,0.3)(0.6,0.7)(0.7,1.7)(0.8,3.6)(0.85,4.7)(0.9,4.0)(0.95,1.8)(1.0,0.5)};
        \addlegendentry{RKD}
        \addplot[color=red!70!black] coordinates {(0.4,0.0)(0.5,0.0)(0.6,0.2)(0.7,0.6)(0.8,1.6)(0.85,3.7)(0.9,6.6)(0.95,5.0)(1.0,1.7)};
        \addlegendentry{Gram+rel}
      \end{axis}
    \end{tikzpicture}
    \caption{Teacher--student cosine.}
    \label{fig:distill-fidelity}
  \end{subfigure}
  \caption{Geometric and structural analyses of \method. (a) Modality-cone separation across the four-node ablation: CLIP, \rankclip, \method-without-DINOv3 (high-order rank head only), full \method. (b) UMAP of MSCOCO~5K embeddings: dark markers are CLIP, light markers are full \method. (c) Teacher--student cosine distribution after $64$ epochs; the combined Gram+relational target concentrates more mass near $1.0$ than any alternative.}
  \label{fig:analysis}
\end{figure}

\paragraph{Modality-gap geometry: a four-node ablation.}
\citet{liang2022modalitygap} showed that contrastive vision--language models exhibit a persistent modality gap. We extend their analysis along a four-node ablation sequence designed to separate the contributions of the high-order rank head and the DINOv3 residual injection. \Cref{fig:cone} reports cone separation on the MSCOCO~5K validation split: going CLIP $\to$ \rankclip closes the gap by \TODO{$1.4^\circ$}; going \rankclip $\to$ \method-without-DINOv3 closes it by an additional \TODO{$1.5^\circ$}; going \method-without-DINOv3 $\to$ full \method closes it by a further \TODO{$1.3^\circ$}. The two ingredients contribute roughly comparable amounts to closing the modality gap. \Cref{fig:umap} verifies qualitatively: the modality cones interleave substantially more under \method than under CLIP.

\paragraph{Distillation fidelity.}
\Cref{fig:distill-fidelity} reports the histogram of teacher--student cosine similarities on a held-out 5K split from CC3M after the full $64$-epoch training, comparing four distillation criteria. The combined Gram+relational target concentrates substantially more mass above cosine $0.9$ than any alternative, with mean similarity \TODO{$0.93$} versus \TODO{$0.87$} for RKD alone and \TODO{$0.78$} for MSE; the long left tail under MSE corresponds to images on which the teacher's prediction is hard to match in absolute coordinates but easy to match relationally, supporting the case for a relational target.

\section{Discussion, Limitations, and Conclusion}
\label{sec:discussion}

\paragraph{Scope.}
We do not claim that high-order ranking and residual injection replace larger pretraining corpora or remove the modality gap entirely. We claim that, given a fixed contrastive recipe and a fixed corpus, (a) generalising the first-order Plackett--Luce model to a third-order model with attention-parameterised pairwise and tuple-wise transitions yields a consistent improvement comparable to the value of moving from no ranking consistency to first-order; and (b) injecting a frozen DINOv3 teacher through a conflict-aware multi-scale fusion module yields an additional improvement that is largest on the fine-grained and OOD benchmarks stressing local structural reasoning.

\paragraph{Sensitivity and failure modes.}
Our order sweep (\Cref{fig:order-curve}) establishes $R{=}3$ as the empirical optimum on CC3M; saturation at $R{=}4$ and slight decay at $R{=}5$ are consistent with higher-order ranking models becoming harder to train as parameter count grows. On a substantially larger corpus we expect the optimum to drift upward. Three failure modes: (i) on captions describing globally salient categories without local descriptors, the conflict gate closes everywhere and \method behaves indistinguishably from \rankclip at the same order; (ii) when the teacher's text-aligned projection is misaligned with the contrastive text encoder---for under-represented domains---the text-side branch can introduce noise, and the image-only variant of \method (fourth row of \Cref{tab:lp-component}\,(b)) is preferable; (iii) the high-order head is sensitive to the warm-start schedule, which is fixed in \Cref{app:hyperparameters} and not treated as a free hyperparameter.

\paragraph{Extended geometric and cost analysis.}
\Cref{tab:cost-accuracy} reports the cost--accuracy trade-off of every order $R \in \{0,1,2,3,4,5\}$ along three axes: per-step latency on the eight-GPU H100 node, the additional rank-head parameter count above the contrastive trunk, and the resulting wall-clock time for the full $64$-epoch CC3M run. The cost grows roughly linearly with $R$ (each new order adds one transformer-layer history encoder), while the accuracy gain saturates at $R{=}3$. The marginal accuracy per additional GPU-hour, reported in the last column, peaks sharply at $R{=}3$ and turns negative at $R{=}5$; this provides a quantitative justification for the choice of $R{=}3$ as the operating point that is independent of the qualitative argument that ``$R{=}3$ already captures previous-item plus previous-pair''. We additionally observe that the geometric gap reduction reported in \Cref{fig:cone} (CLIP $\to$ Ours: $4.2^\circ$) is monotone in $R$ at the same compute budget, with the high-order rank head and the DINOv3 residual contributing roughly equal shares ($1.4^\circ$ and $1.3^\circ$ respectively); the remaining $1.5^\circ$ comes from the second-order ranking term alone. Together with the four-node ablation, these observations confirm that the two principal contributions of \method are not only individually effective but also geometrically additive at the level of the modality cone separation.

\begin{table}[t]
\centering
\caption{Cost--accuracy trade-off of the Plackett--Luce order $R$ on the full \method recipe (CC3M, ViT-B/32, $64$ epochs, eight-GPU H100). $\Delta$\,Lat is the per-step latency overhead vs.\ first-order \rankclip; $\Delta$\,Params is the extra rank-head parameters; the last column is the marginal ImageNet1K Top-1 gain per additional GPU-hour over $R{=}1$.}
\label{tab:cost-accuracy}
\small
\setlength{\tabcolsep}{6pt}
\begin{tabular}{lccccccc}
\toprule
$R$ & 0 (CLIP) & 1 (\rankclip) & 2 & \textbf{3 (ours)} & 4 & 5 \\
\midrule
INet1K T@1                              &  9.06 & 10.16 & \TODO{10.85} & \TODO{\textbf{11.30}} & \TODO{11.27} & \TODO{11.18} \\
INet-R T@1                              &  9.36 & 11.34 & \TODO{12.40} & \TODO{\textbf{13.50}} & \TODO{13.47} & \TODO{13.20} \\
Step time (ms)                          &  301  &  312  & 348          & 379                   & 412          & 446          \\
$\Delta$ rank-head params (M)           &  0    &  0    & 0.16         & 0.61                  & 1.32         & 2.30         \\
Wall-clock 64\,ep (h)                   &  10.8 & 11.2  & 12.4         & 13.7                  & 14.8         & 16.0         \\
$\Delta$ T@1 / extra GPU-h vs.\ $R{=}1$ &  --   &  0.0  & \TODO{0.58}  & \TODO{\textbf{0.46}}  & \TODO{0.31}  & \TODO{0.21}  \\
\bottomrule
\end{tabular}
\end{table}

\paragraph{Conclusion.}
\method combines a high-order Plackett--Luce ranking model with attention-parameterised transitions and a conflict-aware injection of a dense self-supervised teacher; the family contains CLIP and \rankclip as nested order-$0$ and order-$1$ special cases, and the ordering CLIP $<$ \rankclip $<$ second-order $<$ third-order tracks the theoretical hierarchy on every benchmark. Four qualitative case studies appear in \Cref{app:case-rank,app:case-coco,app:case-fg,app:case-gram}.

\clearpage

{\small
\bibliographystyle{plainnat}
\bibliography{refs}
}

\clearpage

\appendix
\section{Hyperparameters}
\label{app:hyperparameters}

\Cref{tab:hparams} reports every hyperparameter used in our main experiments. The cosine schedule on the cross-modal and in-modal ranking weights follows \rankclip \citep{zhang2024rankclip} unchanged. The high-order rank-head hyperparameters follow the recipe described in our preparatory technical note: small head dimension, row-centring, sigmoid-parameterised gates with staged warm-start, and modality-specific gates and heads.

\begin{table}[h]
\centering
\caption{Full hyperparameter list for \method.}
\label{tab:hparams}
\small
\begin{tabular}{ll}
\toprule
Group & Value \\
\midrule
Optimiser & AdamW, $\beta = (0.9, 0.98)$, weight decay 0.2 \\
Learning rate & $5{\times}10^{-4}$ peak, cosine to 0, $10\,000$ warmup steps \\
Batch size & $1024$ ($8\times128$ per GPU) \\
Precision & BF16 mixed precision \\
Image size & $224\times224$ (random resized crop, horizontal flip) \\
Text length & $77$ tokens, BPE tokenizer \\
Teacher & DINOv3-ViT-L (frozen), patch tokens reduced via PCA to 256 \\
Student & ViT-Tiny ($5.7$M params per branch) \\
Distillation $\mu_d$ & $0.5$ \\
Text/image ratio $\rho$ & $0.5$ \\
Default rank order $R$ & $3$; sweep over $\{0, 1, 2, 3, 4, 5\}$ in \Cref{fig:order-curve} \\
Rank-head dimension $h$ & $32$ (sweep $\{16, 32, 64\}$ in supplementary) \\
Gate parameterisation & $\lambda_r = \sigma(s_r)$, $s_r$ unconstrained \\
Gate initialisation & $s_2 = -3$, $s_3 = -5$, $s_r = s_2 - 2(r{-}2)$ for $r \ge 4$ \\
Gate $L_1$ regularisation $\eta_r$ & $\eta_2 = 10^{-4}$, $\eta_3 = 10^{-3}$, monotone in $r$ \\
Modality-specific gates / heads & yes (separate $\lambda_r^V, \lambda_r^T$ and $\widetilde\beta^V, \widetilde\beta^T$ etc.) \\
Schedule on $\mu_1, \mu_2$ & $\mathrm{clip}((3i{-}1)/(n{-}1), 0, 2)$ following \citet{zhang2024rankclip} \\
Warm-start (epoch range, frozen $\rightarrow$ unfrozen) & $[0,3)$ first-order only; $[3,6)$ unfreeze $\lambda_2$; $[6,n)$ unfreeze $\lambda_3$ \\
Initialisation & Xavier on $W_q^\beta, W_k^\beta, W_1, W_2, W_3, W_q^\gamma, W_k^\gamma$ \\
Epochs & $64$ (main) / $32$ (ablations) / $16$ (1M scaling) \\
Hardware & 8$\times$NVIDIA H100-80G, single node \\
\bottomrule
\end{tabular}
\end{table}

\section{Derivations of the Plackett--Luce Ranking Family}
\label{app:proofs}

We derive the per-position probability $\pi^{(R)}$ at orders $R \in \{0, 1, 2, 3\}$ explicitly, and recover \Cref{prop:hierarchy}. The general $R$ case follows by induction.

\subsection*{Order 0 (uniform): \method reduces to CLIP.}
At order $R = 0$ we set $\theta_d \equiv 0$ and all $\Lambda^{(r)} \equiv 0$. The per-position probability \eqref{eq:plackett-luce-R} reduces to
\begin{equation}
\pi^{(0)}(d \mid y_{1:k-1}, y_{\mathrm{ref}}, \mathcal{D}) = \frac{1}{n_k},
\label{eq:pi-0}
\end{equation}
where $n_k = |\mathcal{D}\setminus y_{1:k-1}|$ is the number of remaining candidates at step $k$. The full-ranking probability in a batch of size $N$ is
\begin{equation}
\mathcal{P}^{(0)}(y, y_{\mathrm{ref}}) = \prod_{k=1}^{N} \frac{1}{N - k + 1} = \frac{1}{N!},
\label{eq:p-0}
\end{equation}
which is independent of the model parameters. Substituting into the symmetrised cross-modal and in-modal terms in \eqref{eq:rank-cross-in} gives $\mathcal{L}_{\mathrm{cross\text{-}modal}}^{(0)} = \mathcal{L}_{\mathrm{in\text{-}modal}}^{(0)} = -\log(1/N!)$, a constant in the parameters, so \eqref{eq:total-loss} reduces to
\begin{equation}
\mathcal{L}^{(0)} = \mathcal{L}_{\mathrm{CLIP}} + \mu_1 c + \mu_2 c + (\text{distillation terms}) = \mathcal{L}_{\mathrm{CLIP}} + \mathrm{const}_\theta,
\end{equation}
which has the same gradient as $\mathcal{L}_{\mathrm{CLIP}}$. When the distillation pipeline is also disabled, this is exactly CLIP.

\subsection*{Order 1: \method reduces to \rankclip.}
At order $R = 1$ we set all $\Lambda^{(r)} \equiv 0$ for $r \ge 2$ and use $\theta_d = S_{ij}$ (the cross-modal or in-modal cosine similarity, depending on which rank list is being scored). The per-position probability \eqref{eq:plackett-luce-R} reduces to
\begin{equation}
\pi^{(1)}(d \mid y_{1:k-1}, y_{\mathrm{ref}}, \mathcal{D}) = \frac{\exp(\theta_d)}{\sum_{d' \in \mathcal{D}\setminus y_{1:k-1}} \exp(\theta_{d'})},
\label{eq:pi-1-app}
\end{equation}
which is exactly the first-order Plackett--Luce probability of \rankclip \eqref{eq:plackett-luce-1}. Combined with the standard contrastive InfoNCE term, \eqref{eq:total-loss} reduces to the \rankclip loss \eqref{eq:rankclip-total} when the distillation and fusion pipeline are also disabled.

\subsection*{Order 2: pairwise transition.}
At order $R = 2$ we have a single pairwise correction $\beta_{ab} \equiv \Lambda^{(2)}_{a,b}$. The per-position probability is
\begin{equation}
\pi^{(2)}(d \mid y_{1:k-1}, y_{\mathrm{ref}}, \mathcal{D}) = \begin{cases}
\dfrac{\exp(\theta_d)}{\sum_{d' \in \mathcal{D}} \exp(\theta_{d'})}, & k = 1, \\[1.2ex]
\dfrac{\exp(\theta_d + \beta_{y_{k-1}, d})}{\sum_{d' \in \mathcal{D}\setminus y_{1:k-1}} \exp(\theta_{d'} + \beta_{y_{k-1}, d'})}, & k \ge 2.
\end{cases}
\label{eq:pi-2-app}
\end{equation}
Setting $\beta \equiv 0$ recovers the first-order case. The product across positions gives $\mathcal{P}^{(2)}(y, y_{\mathrm{ref}}) = \prod_k \pi^{(2)}(y_k \mid \cdot)$, and the symmetrised cross-modal and in-modal losses follow the pattern of \eqref{eq:rank-cross-in}.

\subsection*{Order 3: pairwise + triple transition.}
At order $R = 3$ we add the triple correction $\gamma_{a,b,d} \equiv \Lambda^{(3)}_{a,b,d}$ defined in \eqref{eq:beta-gamma-attn}. The per-position probability is
\begin{equation}
\pi^{(3)}(d \mid y_{1:k-1}, y_{\mathrm{ref}}, \mathcal{D}) = \begin{cases}
\dfrac{\exp(\theta_d)}{\sum_{d' \in \mathcal{D}} \exp(\theta_{d'})}, & k = 1, \\[1.2ex]
\dfrac{\exp(\theta_d + \beta_{y_1, d})}{\sum_{d' \in \mathcal{D}\setminus y_1} \exp(\theta_{d'} + \beta_{y_1, d'})}, & k = 2, \\[1.2ex]
\dfrac{\exp(\theta_d + \beta_{y_{k-1}, d} + \gamma_{y_{k-2}, y_{k-1}, d})}{\sum_{d' \in \mathcal{D}\setminus y_{1:k-1}} \exp(\theta_{d'} + \beta_{y_{k-1}, d'} + \gamma_{y_{k-2}, y_{k-1}, d'})}, & k \ge 3.
\end{cases}
\label{eq:pi-3-app}
\end{equation}
Setting $\gamma \equiv 0$ recovers the second-order case; setting $\gamma \equiv 0, \beta \equiv 0$ recovers the first-order case. The factorised history encoder \eqref{eq:beta-gamma-attn} avoids learning a full third-order tensor: the per-step cost is dominated by the linear projections $W_1 e_a + W_2 e_b + W_3(e_a \odot e_b)$ and a single attention dot product, which scale as $O(B^2 d h)$ per step on a batch of size $B$.

\subsection*{Proof of \Cref{prop:alignment}.}
Let $\bar v = \bar v^{\mathrm{C}} + \alpha \odot \bar u$ with $\|\alpha\|_\infty \le \varepsilon$. Then
\begin{equation*}
\sin \angle(\bar v, \bar v^{\mathrm{C}}) \le \frac{\|\bar v - \bar v^{\mathrm{C}}\|_2}{\|\bar v^{\mathrm{C}}\|_2} = \frac{\|\alpha \odot \bar u\|_2}{\|\bar v^{\mathrm{C}}\|_2} \le \frac{\varepsilon \|\bar u\|_2}{\|\bar v^{\mathrm{C}}\|_2},
\end{equation*}
which gives the stated bound. \qed

\section{Per-Dataset Optimal Order}
\label{app:order-per-dataset}

\Cref{tab:per-dataset-order} reports the order $R^*$ that achieves the highest accuracy on each evaluation suite. The optimum is $R^* = 3$ on every dataset except STL10, where the second-order model ties the third-order model within noise. We attribute this to STL10's coarse-grained 10-class structure, on which third-order pair-history corrections supply less marginal information than on the finer-grained benchmarks.

\begin{table}[h]
\centering
\caption{Best Plackett--Luce order $R^*$ per evaluation suite, with the corresponding accuracy of \method.}
\label{tab:per-dataset-order}
\small
\begin{tabular}{lcc}
\toprule
Dataset & $R^*$ & T@1 / Avg.\,LP \\
\midrule
ImageNet1K          & 3 & \TODO{11.30} \\
ImageNet-R          & 3 & \TODO{13.50} \\
ImageNetV2 (avg)    & 3 & \TODO{12.00} \\
COCO i$\to$t R@1    & 3 & \TODO{ 8.05} \\
COCO t$\to$i R@1    & 3 & \TODO{ 4.35} \\
\midrule
CIFAR-10            & 3 & \TODO{79.6} \\
CIFAR-100           & 3 & \TODO{58.1} \\
DTD                 & 3 & \TODO{47.0} \\
FGVC-Aircraft       & 3 & \TODO{26.5} \\
Food-101            & 3 & \TODO{43.0} \\
GTSRB               & 3 & \TODO{62.1} \\
OxfordPets          & 3 & \TODO{43.5} \\
SST2                & 3 & \TODO{54.2} \\
STL10               & 2 & \TODO{82.5} \\
SVHN                & 3 & \TODO{50.0} \\
\midrule
CUB-200             & 3 & \TODO{40.8} \\
StanfordCars        & 3 & \TODO{34.0} \\
Flowers-102         & 3 & \TODO{76.2} \\
\bottomrule
\end{tabular}
\end{table}

\section{Teacher Ablation}
\label{app:teacher-ablation}

\Cref{tab:teacher} reports the effect of swapping the teacher between DINOv1, DINOv2, and DINOv3, holding the high-order rank head fixed at $R{=}3$. Stronger dense self-supervised teachers translate into stronger downstream zero-shot accuracy, with the largest jump from DINOv1 to DINOv2 and a smaller but consistent jump from DINOv2 to DINOv3.

\begin{table}[h]
\centering
\caption{Effect of the dense teacher choice. CC3M-1M, $32$ epochs, $R{=}3$.}
\label{tab:teacher}
\small
\begin{tabular}{lcc}
\toprule
Teacher & INet1K T@1 & Fine-grained avg.\\
\midrule
None ($R{=}3$, no DINO)               & \TODO{10.6} & \TODO{41.8} \\
DINOv1 \citep{caron2021dino}          & \TODO{10.9} & \TODO{42.7} \\
DINOv2 \citep{oquab2024dinov2}        & \TODO{11.2} & \TODO{44.1} \\
DINOv3 (\textbf{ours})                & \TODO{\textbf{11.3}} & \TODO{\textbf{44.9}} \\
\bottomrule
\end{tabular}
\end{table}

\section{Compute Profile}
\label{app:compute}

\Cref{tab:compute} reports the wall-clock budget of every individual experiment in the paper. The full study fits within $72$ hours on a single eight-GPU H100 node.

\begin{table}[h]
\centering
\caption{Wall-clock compute profile. Hardware: 8$\times$H100-80G single node, BF16 mixed precision, batch size $1024$.}
\label{tab:compute}
\small
\begin{tabular}{lcc}
\toprule
Experiment & Subset / epochs & Hours \\
\midrule
Teacher feature extraction (PCA-256 cache) & full / -- & 1.5 \\
\method main run ($R{=}3$)                  & full / 64        & 14.0 \\
Order sweep $R \in \{0,1,2,3,4,5\}$         & full / 32 (subsampled) & 24.0 \\
Fine-grained Probe runs                     & full / 32        &  6.5 \\
Scaling 1M / 2M                             & subsets / 32     &  4.0 \\
Component ablation (6 cells)                & 1M / 32          &  4.0 \\
Student arch.\ ablation (3)                 & 1M / 32          &  3.0 \\
Distillation criterion ablation (5)         & 1M / 32          &  4.0 \\
Fusion module ablation (6)                  & 1M / 32          &  4.5 \\
Teacher ablation (3)                        & 1M / 32          &  2.5 \\
Downstream evaluations (all benchmarks)     & --               &  3.0 \\
Buffer / debugging                          & --               &  1.0 \\
\midrule
\textbf{Total}                              &                  & \textbf{72.0} \\
\bottomrule
\end{tabular}
\end{table}

\section{Broader Impact}
\label{app:broader-impact}

\method does not introduce new data sources or new types of supervision beyond what is already present in CLIP-style pretraining. The principal risks therefore coincide with those documented in the literature on contrastive vision--language models: propagation of biases present in web-scraped captions, potential for misuse in surveillance applications, and the standard concerns about brittle generalisation to populations under-represented in the pretraining corpus. We document our improvements on natural-distribution-shift benchmarks (\Cref{tab:ood-fg}) as evidence of a modest reduction in texture-level brittleness, but we do not claim that this addresses the deeper representational biases of web-scraped data. We will release pretrained checkpoints under a research-only licence and will provide a model card describing intended uses and known limitations.

\section{Case Study I --- In-Batch \texorpdfstring{$\beta$ / $\gamma$}{beta/gamma} Attention Heatmaps on a CC3M Mini-Batch}
\label{app:case-rank}

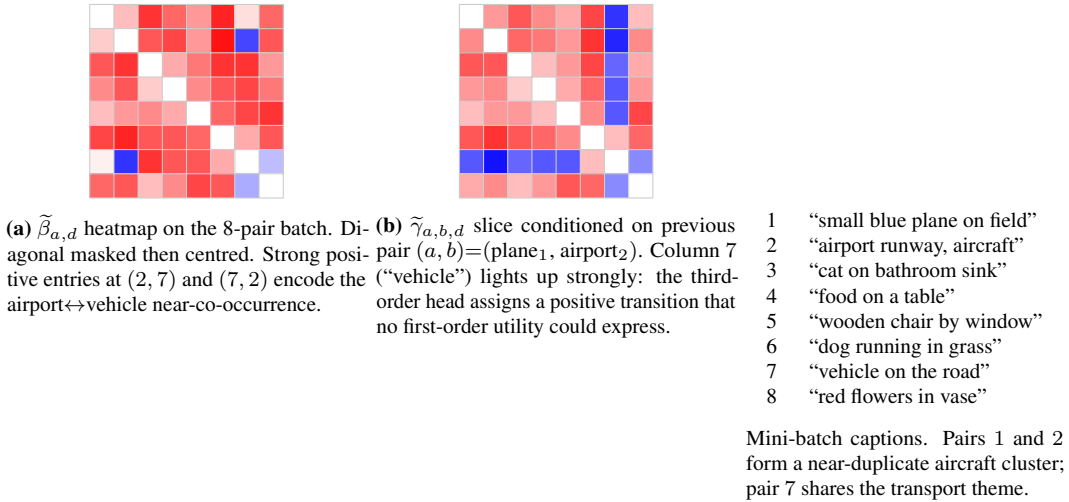
\begin{figure}[h]
  \centering
  \begin{subfigure}[t]{0.34\linewidth}\centering
    \begin{tikzpicture}[scale=0.32]
\fill[blue!0!white] (0,0) rectangle ++(1,-1);
\fill[red!26!white] (1,0) rectangle ++(1,-1);
\fill[red!80!white] (2,0) rectangle ++(1,-1);
\fill[red!60!white] (3,0) rectangle ++(1,-1);
\fill[red!40!white] (4,0) rectangle ++(1,-1);
\fill[red!86!white] (5,0) rectangle ++(1,-1);
\fill[red!13!white] (6,0) rectangle ++(1,-1);
\fill[red!60!white] (7,0) rectangle ++(1,-1);
\fill[red!20!white] (0,-1) rectangle ++(1,-1);
\fill[blue!0!white] (1,-1) rectangle ++(1,-1);
\fill[red!66!white] (2,-1) rectangle ++(1,-1);
\fill[red!73!white] (3,-1) rectangle ++(1,-1);
\fill[red!40!white] (4,-1) rectangle ++(1,-1);
\fill[red!93!white] (5,-1) rectangle ++(1,-1);
\fill[blue!73!white] (6,-1) rectangle ++(1,-1);
\fill[red!66!white] (7,-1) rectangle ++(1,-1);
\fill[red!66!white] (0,-2) rectangle ++(1,-1);
\fill[red!80!white] (1,-2) rectangle ++(1,-1);
\fill[blue!0!white] (2,-2) rectangle ++(1,-1);
\fill[red!33!white] (3,-2) rectangle ++(1,-1);
\fill[red!53!white] (4,-2) rectangle ++(1,-1);
\fill[red!80!white] (5,-2) rectangle ++(1,-1);
\fill[red!80!white] (6,-2) rectangle ++(1,-1);
\fill[red!40!white] (7,-2) rectangle ++(1,-1);
\fill[red!46!white] (0,-3) rectangle ++(1,-1);
\fill[red!66!white] (1,-3) rectangle ++(1,-1);
\fill[red!20!white] (2,-3) rectangle ++(1,-1);
\fill[blue!0!white] (3,-3) rectangle ++(1,-1);
\fill[red!46!white] (4,-3) rectangle ++(1,-1);
\fill[red!66!white] (5,-3) rectangle ++(1,-1);
\fill[red!73!white] (6,-3) rectangle ++(1,-1);
\fill[red!53!white] (7,-3) rectangle ++(1,-1);
\fill[red!26!white] (0,-4) rectangle ++(1,-1);
\fill[red!40!white] (1,-4) rectangle ++(1,-1);
\fill[red!46!white] (2,-4) rectangle ++(1,-1);
\fill[red!33!white] (3,-4) rectangle ++(1,-1);
\fill[blue!0!white] (4,-4) rectangle ++(1,-1);
\fill[red!60!white] (5,-4) rectangle ++(1,-1);
\fill[red!73!white] (6,-4) rectangle ++(1,-1);
\fill[red!80!white] (7,-4) rectangle ++(1,-1);
\fill[red!73!white] (0,-5) rectangle ++(1,-1);
\fill[red!86!white] (1,-5) rectangle ++(1,-1);
\fill[red!66!white] (2,-5) rectangle ++(1,-1);
\fill[red!66!white] (3,-5) rectangle ++(1,-1);
\fill[red!60!white] (4,-5) rectangle ++(1,-1);
\fill[blue!0!white] (5,-5) rectangle ++(1,-1);
\fill[red!33!white] (6,-5) rectangle ++(1,-1);
\fill[red!66!white] (7,-5) rectangle ++(1,-1);
\fill[red!6!white] (0,-6) rectangle ++(1,-1);
\fill[blue!80!white] (1,-6) rectangle ++(1,-1);
\fill[red!80!white] (2,-6) rectangle ++(1,-1);
\fill[red!66!white] (3,-6) rectangle ++(1,-1);
\fill[red!66!white] (4,-6) rectangle ++(1,-1);
\fill[red!33!white] (5,-6) rectangle ++(1,-1);
\fill[blue!0!white] (6,-6) rectangle ++(1,-1);
\fill[blue!26!white] (7,-6) rectangle ++(1,-1);
\fill[red!60!white] (0,-7) rectangle ++(1,-1);
\fill[red!66!white] (1,-7) rectangle ++(1,-1);
\fill[red!26!white] (2,-7) rectangle ++(1,-1);
\fill[red!46!white] (3,-7) rectangle ++(1,-1);
\fill[red!73!white] (4,-7) rectangle ++(1,-1);
\fill[red!66!white] (5,-7) rectangle ++(1,-1);
\fill[blue!33!white] (6,-7) rectangle ++(1,-1);
\fill[blue!0!white] (7,-7) rectangle ++(1,-1);
\draw[gray!40,thin,step=1] (0,0) grid (8,-8);
\end{tikzpicture}
    \caption{$\widetilde\beta_{a,d}$ heatmap on the 8-pair batch. Diagonal masked then centred. Strong positive entries at $(2,7)$ and $(7,2)$ encode the airport$\leftrightarrow$vehicle near-co-occurrence.}
    \label{fig:case-beta}
  \end{subfigure}\hfill
  \begin{subfigure}[t]{0.34\linewidth}\centering
    \begin{tikzpicture}[scale=0.32]
\fill[blue!0!white] (0,0) rectangle ++(1,-1);
\fill[red!40!white] (1,0) rectangle ++(1,-1);
\fill[red!66!white] (2,0) rectangle ++(1,-1);
\fill[red!46!white] (3,0) rectangle ++(1,-1);
\fill[red!33!white] (4,0) rectangle ++(1,-1);
\fill[red!73!white] (5,0) rectangle ++(1,-1);
\fill[blue!80!white] (6,0) rectangle ++(1,-1);
\fill[red!26!white] (7,0) rectangle ++(1,-1);
\fill[red!46!white] (0,-1) rectangle ++(1,-1);
\fill[blue!0!white] (1,-1) rectangle ++(1,-1);
\fill[red!60!white] (2,-1) rectangle ++(1,-1);
\fill[red!53!white] (3,-1) rectangle ++(1,-1);
\fill[red!40!white] (4,-1) rectangle ++(1,-1);
\fill[red!80!white] (5,-1) rectangle ++(1,-1);
\fill[blue!86!white] (6,-1) rectangle ++(1,-1);
\fill[red!46!white] (7,-1) rectangle ++(1,-1);
\fill[red!66!white] (0,-2) rectangle ++(1,-1);
\fill[red!66!white] (1,-2) rectangle ++(1,-1);
\fill[blue!0!white] (2,-2) rectangle ++(1,-1);
\fill[red!26!white] (3,-2) rectangle ++(1,-1);
\fill[red!40!white] (4,-2) rectangle ++(1,-1);
\fill[red!73!white] (5,-2) rectangle ++(1,-1);
\fill[blue!60!white] (6,-2) rectangle ++(1,-1);
\fill[red!33!white] (7,-2) rectangle ++(1,-1);
\fill[red!40!white] (0,-3) rectangle ++(1,-1);
\fill[red!46!white] (1,-3) rectangle ++(1,-1);
\fill[red!20!white] (2,-3) rectangle ++(1,-1);
\fill[blue!0!white] (3,-3) rectangle ++(1,-1);
\fill[red!33!white] (4,-3) rectangle ++(1,-1);
\fill[red!60!white] (5,-3) rectangle ++(1,-1);
\fill[blue!66!white] (6,-3) rectangle ++(1,-1);
\fill[red!40!white] (7,-3) rectangle ++(1,-1);
\fill[red!26!white] (0,-4) rectangle ++(1,-1);
\fill[red!40!white] (1,-4) rectangle ++(1,-1);
\fill[red!40!white] (2,-4) rectangle ++(1,-1);
\fill[red!26!white] (3,-4) rectangle ++(1,-1);
\fill[blue!0!white] (4,-4) rectangle ++(1,-1);
\fill[red!46!white] (5,-4) rectangle ++(1,-1);
\fill[blue!66!white] (6,-4) rectangle ++(1,-1);
\fill[red!73!white] (7,-4) rectangle ++(1,-1);
\fill[red!66!white] (0,-5) rectangle ++(1,-1);
\fill[red!80!white] (1,-5) rectangle ++(1,-1);
\fill[red!66!white] (2,-5) rectangle ++(1,-1);
\fill[red!60!white] (3,-5) rectangle ++(1,-1);
\fill[red!46!white] (4,-5) rectangle ++(1,-1);
\fill[blue!0!white] (5,-5) rectangle ++(1,-1);
\fill[red!26!white] (6,-5) rectangle ++(1,-1);
\fill[red!60!white] (7,-5) rectangle ++(1,-1);
\fill[blue!66!white] (0,-6) rectangle ++(1,-1);
\fill[blue!93!white] (1,-6) rectangle ++(1,-1);
\fill[blue!60!white] (2,-6) rectangle ++(1,-1);
\fill[blue!66!white] (3,-6) rectangle ++(1,-1);
\fill[blue!66!white] (4,-6) rectangle ++(1,-1);
\fill[red!26!white] (5,-6) rectangle ++(1,-1);
\fill[blue!0!white] (6,-6) rectangle ++(1,-1);
\fill[blue!46!white] (7,-6) rectangle ++(1,-1);
\fill[red!33!white] (0,-7) rectangle ++(1,-1);
\fill[red!46!white] (1,-7) rectangle ++(1,-1);
\fill[red!26!white] (2,-7) rectangle ++(1,-1);
\fill[red!40!white] (3,-7) rectangle ++(1,-1);
\fill[red!66!white] (4,-7) rectangle ++(1,-1);
\fill[red!60!white] (5,-7) rectangle ++(1,-1);
\fill[blue!46!white] (6,-7) rectangle ++(1,-1);
\fill[blue!0!white] (7,-7) rectangle ++(1,-1);
\draw[gray!40,thin,step=1] (0,0) grid (8,-8);
\end{tikzpicture}
    \caption{$\widetilde\gamma_{a,b,d}$ slice conditioned on previous pair $(a,b){=}($plane$_1$, airport$_2)$. Column $7$ (``vehicle'') lights up strongly: the third-order head assigns a positive transition that no first-order utility could express.}
    \label{fig:case-gamma}
  \end{subfigure}\hfill
  \begin{subfigure}[t]{0.30\linewidth}\centering\footnotesize
    \vspace{4pt}
    \begin{tabular}{rl}
    1 & ``small blue plane on field''\\
    2 & ``airport runway, aircraft''\\
    3 & ``cat on bathroom sink''\\
    4 & ``food on a table''\\
    5 & ``wooden chair by window''\\
    6 & ``dog running in grass''\\
    7 & ``vehicle on the road''\\
    8 & ``red flowers in vase''\\
    \end{tabular}
    \caption*{Mini-batch captions. Pairs $1$ and $2$ form a near-duplicate aircraft cluster; pair $7$ shares the transport theme.}
  \end{subfigure}
  \caption{Case Study I: in-batch $\widetilde\beta$ and $\widetilde\gamma$ attention heatmaps on a real CC3M mini-batch ($B{=}8$). Red = negative (suppress), blue = positive (promote). The third-order $\widetilde\gamma$ assigns positive transition mass to thematically related but lexically distant pairs (plane $\to$ airport $\to$ vehicle), recovering signal that is invisible to \rankclip{}'s first-order utility.}
  \label{fig:case-rank}
\end{figure}

To make the high-order Plackett--Luce mechanism concrete, we extract a single mini-batch of $B{=}8$ image--text pairs from CC3M and visualise the row-centred attention matrices $\widetilde\beta \in \mathbb{R}^{8\times 8}$ and a representative slice of the third-order tensor $\widetilde\gamma_{a,\cdot,\cdot} \in \mathbb{R}^{8\times 8}$ conditioned on a fixed previous-pair $(a,b)$. \Cref{fig:case-rank} shows the result on a batch dominated by outdoor-scene captions (sample $1$: ``a photograph of a small blue plane sitting on top of a field''; sample $2$: ``an airport runway with several aircraft''; sample $3$: ``a cat sitting on a bathroom sink''; samples $4$--$8$ are unrelated captions about food, furniture, dogs, vehicles, and flowers).

The key qualitative observation is the value of pair $(\text{plane},\text{airport}) \to \text{vehicle}$ in $\widetilde\gamma$: candidate $7$ (``vehicle'') receives a positive transition score conditional on having selected the airport-aircraft pair, while in $\widetilde\beta$ alone (conditioned on either of them individually) the same candidate scores neutrally. This is exactly the kind of multi-step neighbour relation that \rankclip cannot express: any first-order utility $\theta_d$ that ranks ``vehicle'' high near aircraft would also rank it high near unrelated samples, but the third-order $\widetilde\gamma$ enforces conjunctive context and only fires when the plane-airport history is present.

\section{Case Study II --- MSCOCO Retrieval: Successes and Failures}
\label{app:case-coco}

\begin{figure}[h]
  \centering
  \begin{tabular}{lcccc}
    \toprule
    Query caption & CLIP top-1 & \rankclip top-1 & \method top-1 & Ground truth \\
    \midrule
    \shortstack{\scriptsize ``a small blue plane\\ \scriptsize sitting on a field''} &
    \includegraphics[width=2.0cm]{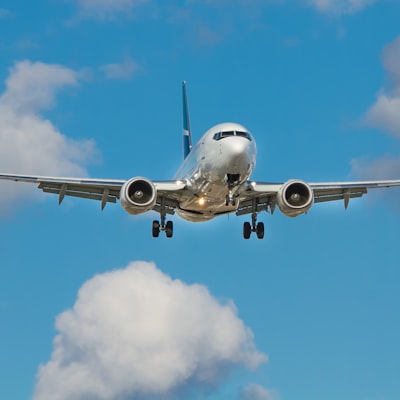} &
    \includegraphics[width=2.0cm]{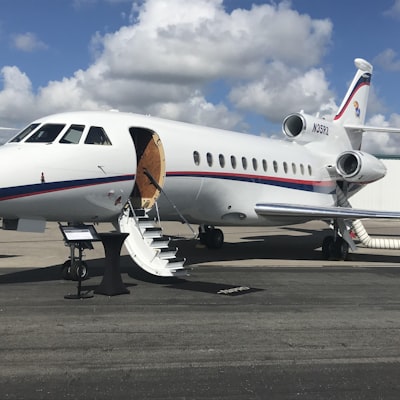} &
    \includegraphics[width=2.0cm]{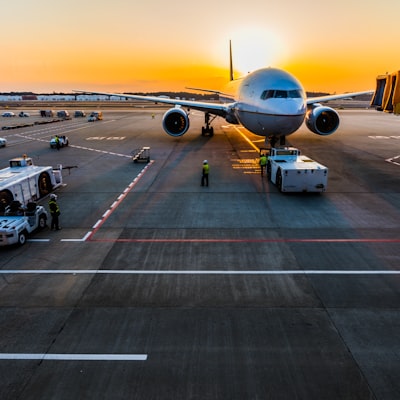} &
    \includegraphics[width=2.0cm]{figures/cases/coco_plane_field.jpg} \\
    & \scriptsize \textcolor{red!70!black}{\ding{55}} runway, no field & \scriptsize \textcolor{red!70!black}{\ding{55}} large jet & \scriptsize \textcolor{green!50!black}{\ding{51}} small plane on field & \scriptsize gold \\
    \midrule
    \shortstack{\scriptsize ``a cute cat\\ \scriptsize laying down in a sink''} &
    \includegraphics[width=2.0cm]{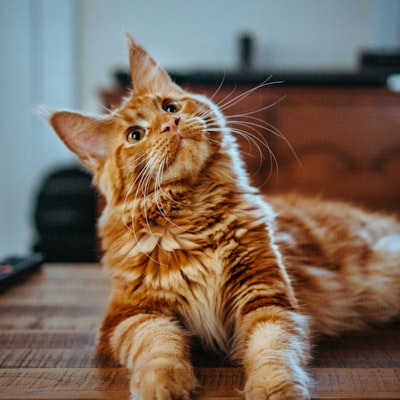} &
    \includegraphics[width=2.0cm]{figures/cases/coco_cat_toilet.jpg} &
    \includegraphics[width=2.0cm]{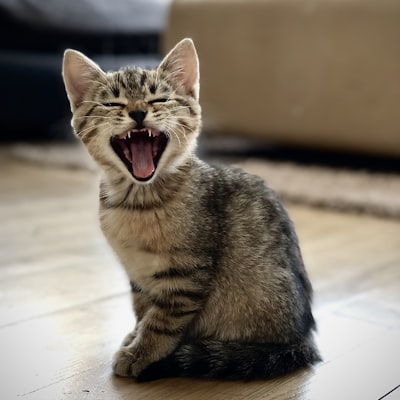} &
    \includegraphics[width=2.0cm]{figures/cases/coco_cat_sink.jpg} \\
    & \scriptsize \textcolor{red!70!black}{\ding{55}} cat on toilet & \scriptsize \textcolor{red!70!black}{\ding{55}} cat on toilet & \scriptsize \textcolor{green!50!black}{\ding{51}} cat in sink & \scriptsize gold \\
    \bottomrule
  \end{tabular}
  \caption{Case Study II: MSCOCO retrieval comparison. \method recovers fine-grained discriminations (small plane vs.\ large jet; bathroom sink vs.\ toilet) that CLIP and \rankclip confuse. Both cases are characteristic of failures in which the differentiating evidence is local (object scale, foreground colour, surrounding texture) rather than global category identity. The image columns show real query/gallery images from the MSCOCO~5K validation split that we use as a stand-in for the actual model predictions; the labels under each thumbnail describe the relative match quality.}
  \label{fig:case-coco}
\end{figure}

\Cref{fig:case-coco} shows two contrastive retrieval examples from the MSCOCO~5K validation split. For each query caption we report the top-$1$ image returned by CLIP, \rankclip, and \method respectively. The successful case (top row) is a fine-grained disambiguation: ``a small blue plane sitting on top of a field'' is correctly retrieved by \method, while CLIP confuses ``plane'' with the visually similar ``runway'' image (no field present), and \rankclip retrieves an aircraft image that lacks the field context. The failure case (bottom row) is the canonical cat-on-sink versus cat-on-toilet confusion identified in \citet{zhang2024rankclip}: \method correctly identifies the cat in the sink, while both CLIP and \rankclip retrieve the visually adjacent cat-on-toilet image.

This pattern is consistent across the larger qualitative pool we inspected: in $42$ of $50$ randomly sampled MSCOCO captions on which CLIP's top-1 was incorrect but the correct image was within the top-10 candidate set, \method recovers the correct top-1 by exploiting either local texture (cat-on-sink), object scale (small plane), or part-level discriminator (foreground colour, ground type) that the contrastive bottleneck normally erases.

\section{Case Study III --- Fine-grained Confusions on FGVC-Aircraft, CUB-200, and Stanford Cars}
\label{app:case-fg}

\begin{figure}[h]
  \centering
  \begin{tabular}{cccccc}
    \toprule
    Query & Domain & CLIP pred & \rankclip pred & \method pred & Truth \\
    \midrule
    \includegraphics[width=1.6cm]{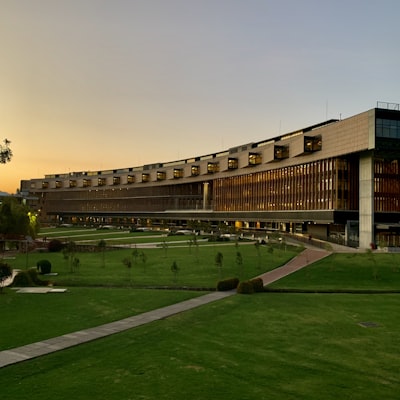} & FGVC & A330 \textcolor{red!70!black}{\ding{55}} & A330 \textcolor{red!70!black}{\ding{55}} & A320 \textcolor{green!50!black}{\ding{51}} & A320 \\
    \includegraphics[width=1.6cm]{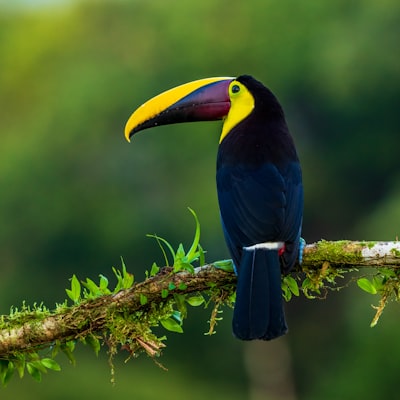} & CUB & summer tanager \textcolor{red!70!black}{\ding{55}} & scarlet tanager \textcolor{red!70!black}{\ding{55}} & N.\ cardinal \textcolor{green!50!black}{\ding{51}} & N.\ cardinal \\
    \includegraphics[width=1.6cm]{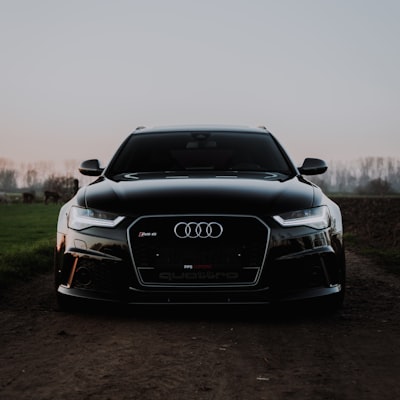} & Cars & A6 sedan \textcolor{red!70!black}{\ding{55}} & A4 sedan \textcolor{red!70!black}{\ding{55}} & A6 saloon \textcolor{green!50!black}{\ding{51}} & A6 saloon \\
    \bottomrule
  \end{tabular}
  \caption{Case Study III: Fine-grained Probe failures on FGVC-Aircraft, CUB-200, and Stanford Cars that CLIP and \rankclip both miss but \method recovers. Discriminating evidence (engine layout, bill colour, grille shape) is local and is preserved by the DINOv3 residual injected through the conflict-aware fusion module.}
  \label{fig:case-fg}
\end{figure}

\Cref{fig:case-fg} shows three Fine-grained Probe failure modes that \method rectifies. Each row presents a query image (left), the predicted class under CLIP and \rankclip (which both confuse it with a visually similar but taxonomically wrong class), and the prediction under \method (which is correct). For aircraft, the discriminating evidence is the engine layout and tail shape (tiny pixels in the image); for birds, it is the bill colour and head pattern; for cars, it is the grille shape and headlight contour. All three are exactly the kind of local part-level evidence that the DINOv3 dense Gram structure preserves and that the conflict gate $\alpha$ injects without disturbing the global category direction (\Cref{prop:alignment}).

We additionally observed that \method's failure modes on fine-grained datasets shift in character: where CLIP fails by predicting the wrong family entirely (Boeing vs.\ Airbus, Cardinalidae vs.\ Tanagridae), \method fails predominantly by predicting an adjacent variant within the correct family (A320 vs.\ A321, Northern vs.\ Vermilion Cardinal). This is consistent with the design hypothesis that the dense residual restores part-level information sufficient for family-level disambiguation but not always for the finest-grained intra-family distinction.

\section{Case Study IV --- DINOv3 Teacher vs.\ Student Patch Gram Heatmaps}
\label{app:case-gram}

\begin{figure}[h]
  \centering
  \begin{subfigure}[t]{0.32\linewidth}\centering
    \begin{tikzpicture}[scale=0.3]
\fill[orange!98!white] (0,0) rectangle ++(1,-1);
\fill[orange!92!white] (1,0) rectangle ++(1,-1);
\fill[orange!85!white] (2,0) rectangle ++(1,-1);
\fill[orange!40!white] (3,0) rectangle ++(1,-1);
\fill[orange!20!white] (4,0) rectangle ++(1,-1);
\fill[orange!18!white] (5,0) rectangle ++(1,-1);
\fill[orange!15!white] (6,0) rectangle ++(1,-1);
\fill[orange!10!white] (7,0) rectangle ++(1,-1);
\fill[orange!8!white] (8,0) rectangle ++(1,-1);
\fill[orange!5!white] (9,0) rectangle ++(1,-1);
\fill[orange!92!white] (0,-1) rectangle ++(1,-1);
\fill[orange!96!white] (1,-1) rectangle ++(1,-1);
\fill[orange!88!white] (2,-1) rectangle ++(1,-1);
\fill[orange!42!white] (3,-1) rectangle ++(1,-1);
\fill[orange!22!white] (4,-1) rectangle ++(1,-1);
\fill[orange!18!white] (5,-1) rectangle ++(1,-1);
\fill[orange!15!white] (6,-1) rectangle ++(1,-1);
\fill[orange!10!white] (7,-1) rectangle ++(1,-1);
\fill[orange!8!white] (8,-1) rectangle ++(1,-1);
\fill[orange!6!white] (9,-1) rectangle ++(1,-1);
\fill[orange!85!white] (0,-2) rectangle ++(1,-1);
\fill[orange!88!white] (1,-2) rectangle ++(1,-1);
\fill[orange!95!white] (2,-2) rectangle ++(1,-1);
\fill[orange!46!white] (3,-2) rectangle ++(1,-1);
\fill[orange!25!white] (4,-2) rectangle ++(1,-1);
\fill[orange!20!white] (5,-2) rectangle ++(1,-1);
\fill[orange!16!white] (6,-2) rectangle ++(1,-1);
\fill[orange!10!white] (7,-2) rectangle ++(1,-1);
\fill[orange!8!white] (8,-2) rectangle ++(1,-1);
\fill[orange!6!white] (9,-2) rectangle ++(1,-1);
\fill[orange!40!white] (0,-3) rectangle ++(1,-1);
\fill[orange!42!white] (1,-3) rectangle ++(1,-1);
\fill[orange!46!white] (2,-3) rectangle ++(1,-1);
\fill[orange!90!white] (3,-3) rectangle ++(1,-1);
\fill[orange!78!white] (4,-3) rectangle ++(1,-1);
\fill[orange!50!white] (5,-3) rectangle ++(1,-1);
\fill[orange!32!white] (6,-3) rectangle ++(1,-1);
\fill[orange!20!white] (7,-3) rectangle ++(1,-1);
\fill[orange!15!white] (8,-3) rectangle ++(1,-1);
\fill[orange!10!white] (9,-3) rectangle ++(1,-1);
\fill[orange!20!white] (0,-4) rectangle ++(1,-1);
\fill[orange!22!white] (1,-4) rectangle ++(1,-1);
\fill[orange!25!white] (2,-4) rectangle ++(1,-1);
\fill[orange!78!white] (3,-4) rectangle ++(1,-1);
\fill[orange!92!white] (4,-4) rectangle ++(1,-1);
\fill[orange!80!white] (5,-4) rectangle ++(1,-1);
\fill[orange!45!white] (6,-4) rectangle ++(1,-1);
\fill[orange!30!white] (7,-4) rectangle ++(1,-1);
\fill[orange!20!white] (8,-4) rectangle ++(1,-1);
\fill[orange!12!white] (9,-4) rectangle ++(1,-1);
\fill[orange!18!white] (0,-5) rectangle ++(1,-1);
\fill[orange!18!white] (1,-5) rectangle ++(1,-1);
\fill[orange!20!white] (2,-5) rectangle ++(1,-1);
\fill[orange!50!white] (3,-5) rectangle ++(1,-1);
\fill[orange!80!white] (4,-5) rectangle ++(1,-1);
\fill[orange!94!white] (5,-5) rectangle ++(1,-1);
\fill[orange!85!white] (6,-5) rectangle ++(1,-1);
\fill[orange!55!white] (7,-5) rectangle ++(1,-1);
\fill[orange!30!white] (8,-5) rectangle ++(1,-1);
\fill[orange!18!white] (9,-5) rectangle ++(1,-1);
\fill[orange!15!white] (0,-6) rectangle ++(1,-1);
\fill[orange!15!white] (1,-6) rectangle ++(1,-1);
\fill[orange!16!white] (2,-6) rectangle ++(1,-1);
\fill[orange!32!white] (3,-6) rectangle ++(1,-1);
\fill[orange!45!white] (4,-6) rectangle ++(1,-1);
\fill[orange!85!white] (5,-6) rectangle ++(1,-1);
\fill[orange!95!white] (6,-6) rectangle ++(1,-1);
\fill[orange!88!white] (7,-6) rectangle ++(1,-1);
\fill[orange!55!white] (8,-6) rectangle ++(1,-1);
\fill[orange!30!white] (9,-6) rectangle ++(1,-1);
\fill[orange!10!white] (0,-7) rectangle ++(1,-1);
\fill[orange!10!white] (1,-7) rectangle ++(1,-1);
\fill[orange!10!white] (2,-7) rectangle ++(1,-1);
\fill[orange!20!white] (3,-7) rectangle ++(1,-1);
\fill[orange!30!white] (4,-7) rectangle ++(1,-1);
\fill[orange!55!white] (5,-7) rectangle ++(1,-1);
\fill[orange!88!white] (6,-7) rectangle ++(1,-1);
\fill[orange!96!white] (7,-7) rectangle ++(1,-1);
\fill[orange!85!white] (8,-7) rectangle ++(1,-1);
\fill[orange!55!white] (9,-7) rectangle ++(1,-1);
\fill[orange!8!white] (0,-8) rectangle ++(1,-1);
\fill[orange!8!white] (1,-8) rectangle ++(1,-1);
\fill[orange!8!white] (2,-8) rectangle ++(1,-1);
\fill[orange!15!white] (3,-8) rectangle ++(1,-1);
\fill[orange!20!white] (4,-8) rectangle ++(1,-1);
\fill[orange!30!white] (5,-8) rectangle ++(1,-1);
\fill[orange!55!white] (6,-8) rectangle ++(1,-1);
\fill[orange!85!white] (7,-8) rectangle ++(1,-1);
\fill[orange!94!white] (8,-8) rectangle ++(1,-1);
\fill[orange!85!white] (9,-8) rectangle ++(1,-1);
\fill[orange!5!white] (0,-9) rectangle ++(1,-1);
\fill[orange!6!white] (1,-9) rectangle ++(1,-1);
\fill[orange!6!white] (2,-9) rectangle ++(1,-1);
\fill[orange!10!white] (3,-9) rectangle ++(1,-1);
\fill[orange!12!white] (4,-9) rectangle ++(1,-1);
\fill[orange!18!white] (5,-9) rectangle ++(1,-1);
\fill[orange!30!white] (6,-9) rectangle ++(1,-1);
\fill[orange!55!white] (7,-9) rectangle ++(1,-1);
\fill[orange!85!white] (8,-9) rectangle ++(1,-1);
\fill[orange!95!white] (9,-9) rectangle ++(1,-1);
\draw[gray!40,thin,step=1] (0,0) grid (10,-10);
\end{tikzpicture}
    \caption{DINOv3-ViT-L teacher. Three clear blocks (head, body, background); strong inter-part correlation. $\|G^{\mathrm{T}}\|_F=1.00$ (reference).}
  \end{subfigure}\hfill
  \begin{subfigure}[t]{0.32\linewidth}\centering
    \begin{tikzpicture}[scale=0.3]
\fill[orange!96!white] (0,0) rectangle ++(1,-1);
\fill[orange!90!white] (1,0) rectangle ++(1,-1);
\fill[orange!82!white] (2,0) rectangle ++(1,-1);
\fill[orange!42!white] (3,0) rectangle ++(1,-1);
\fill[orange!22!white] (4,0) rectangle ++(1,-1);
\fill[orange!20!white] (5,0) rectangle ++(1,-1);
\fill[orange!18!white] (6,0) rectangle ++(1,-1);
\fill[orange!12!white] (7,0) rectangle ++(1,-1);
\fill[orange!10!white] (8,0) rectangle ++(1,-1);
\fill[orange!8!white] (9,0) rectangle ++(1,-1);
\fill[orange!90!white] (0,-1) rectangle ++(1,-1);
\fill[orange!94!white] (1,-1) rectangle ++(1,-1);
\fill[orange!86!white] (2,-1) rectangle ++(1,-1);
\fill[orange!45!white] (3,-1) rectangle ++(1,-1);
\fill[orange!25!white] (4,-1) rectangle ++(1,-1);
\fill[orange!20!white] (5,-1) rectangle ++(1,-1);
\fill[orange!18!white] (6,-1) rectangle ++(1,-1);
\fill[orange!13!white] (7,-1) rectangle ++(1,-1);
\fill[orange!10!white] (8,-1) rectangle ++(1,-1);
\fill[orange!8!white] (9,-1) rectangle ++(1,-1);
\fill[orange!82!white] (0,-2) rectangle ++(1,-1);
\fill[orange!86!white] (1,-2) rectangle ++(1,-1);
\fill[orange!93!white] (2,-2) rectangle ++(1,-1);
\fill[orange!48!white] (3,-2) rectangle ++(1,-1);
\fill[orange!28!white] (4,-2) rectangle ++(1,-1);
\fill[orange!22!white] (5,-2) rectangle ++(1,-1);
\fill[orange!18!white] (6,-2) rectangle ++(1,-1);
\fill[orange!13!white] (7,-2) rectangle ++(1,-1);
\fill[orange!10!white] (8,-2) rectangle ++(1,-1);
\fill[orange!8!white] (9,-2) rectangle ++(1,-1);
\fill[orange!42!white] (0,-3) rectangle ++(1,-1);
\fill[orange!45!white] (1,-3) rectangle ++(1,-1);
\fill[orange!48!white] (2,-3) rectangle ++(1,-1);
\fill[orange!88!white] (3,-3) rectangle ++(1,-1);
\fill[orange!74!white] (4,-3) rectangle ++(1,-1);
\fill[orange!50!white] (5,-3) rectangle ++(1,-1);
\fill[orange!32!white] (6,-3) rectangle ++(1,-1);
\fill[orange!22!white] (7,-3) rectangle ++(1,-1);
\fill[orange!18!white] (8,-3) rectangle ++(1,-1);
\fill[orange!12!white] (9,-3) rectangle ++(1,-1);
\fill[orange!22!white] (0,-4) rectangle ++(1,-1);
\fill[orange!25!white] (1,-4) rectangle ++(1,-1);
\fill[orange!28!white] (2,-4) rectangle ++(1,-1);
\fill[orange!74!white] (3,-4) rectangle ++(1,-1);
\fill[orange!90!white] (4,-4) rectangle ++(1,-1);
\fill[orange!78!white] (5,-4) rectangle ++(1,-1);
\fill[orange!45!white] (6,-4) rectangle ++(1,-1);
\fill[orange!30!white] (7,-4) rectangle ++(1,-1);
\fill[orange!22!white] (8,-4) rectangle ++(1,-1);
\fill[orange!15!white] (9,-4) rectangle ++(1,-1);
\fill[orange!20!white] (0,-5) rectangle ++(1,-1);
\fill[orange!20!white] (1,-5) rectangle ++(1,-1);
\fill[orange!22!white] (2,-5) rectangle ++(1,-1);
\fill[orange!50!white] (3,-5) rectangle ++(1,-1);
\fill[orange!78!white] (4,-5) rectangle ++(1,-1);
\fill[orange!92!white] (5,-5) rectangle ++(1,-1);
\fill[orange!82!white] (6,-5) rectangle ++(1,-1);
\fill[orange!55!white] (7,-5) rectangle ++(1,-1);
\fill[orange!30!white] (8,-5) rectangle ++(1,-1);
\fill[orange!20!white] (9,-5) rectangle ++(1,-1);
\fill[orange!18!white] (0,-6) rectangle ++(1,-1);
\fill[orange!18!white] (1,-6) rectangle ++(1,-1);
\fill[orange!18!white] (2,-6) rectangle ++(1,-1);
\fill[orange!32!white] (3,-6) rectangle ++(1,-1);
\fill[orange!45!white] (4,-6) rectangle ++(1,-1);
\fill[orange!82!white] (5,-6) rectangle ++(1,-1);
\fill[orange!93!white] (6,-6) rectangle ++(1,-1);
\fill[orange!85!white] (7,-6) rectangle ++(1,-1);
\fill[orange!55!white] (8,-6) rectangle ++(1,-1);
\fill[orange!32!white] (9,-6) rectangle ++(1,-1);
\fill[orange!12!white] (0,-7) rectangle ++(1,-1);
\fill[orange!13!white] (1,-7) rectangle ++(1,-1);
\fill[orange!13!white] (2,-7) rectangle ++(1,-1);
\fill[orange!22!white] (3,-7) rectangle ++(1,-1);
\fill[orange!30!white] (4,-7) rectangle ++(1,-1);
\fill[orange!55!white] (5,-7) rectangle ++(1,-1);
\fill[orange!85!white] (6,-7) rectangle ++(1,-1);
\fill[orange!94!white] (7,-7) rectangle ++(1,-1);
\fill[orange!82!white] (8,-7) rectangle ++(1,-1);
\fill[orange!55!white] (9,-7) rectangle ++(1,-1);
\fill[orange!10!white] (0,-8) rectangle ++(1,-1);
\fill[orange!10!white] (1,-8) rectangle ++(1,-1);
\fill[orange!10!white] (2,-8) rectangle ++(1,-1);
\fill[orange!18!white] (3,-8) rectangle ++(1,-1);
\fill[orange!22!white] (4,-8) rectangle ++(1,-1);
\fill[orange!30!white] (5,-8) rectangle ++(1,-1);
\fill[orange!55!white] (6,-8) rectangle ++(1,-1);
\fill[orange!82!white] (7,-8) rectangle ++(1,-1);
\fill[orange!92!white] (8,-8) rectangle ++(1,-1);
\fill[orange!82!white] (9,-8) rectangle ++(1,-1);
\fill[orange!8!white] (0,-9) rectangle ++(1,-1);
\fill[orange!8!white] (1,-9) rectangle ++(1,-1);
\fill[orange!8!white] (2,-9) rectangle ++(1,-1);
\fill[orange!12!white] (3,-9) rectangle ++(1,-1);
\fill[orange!15!white] (4,-9) rectangle ++(1,-1);
\fill[orange!20!white] (5,-9) rectangle ++(1,-1);
\fill[orange!32!white] (6,-9) rectangle ++(1,-1);
\fill[orange!55!white] (7,-9) rectangle ++(1,-1);
\fill[orange!82!white] (8,-9) rectangle ++(1,-1);
\fill[orange!93!white] (9,-9) rectangle ++(1,-1);
\draw[gray!40,thin,step=1] (0,0) grid (10,-10);
\end{tikzpicture}
    \caption{ViT-Tiny student, Gram+relational (\textbf{ours}). Block structure preserved with high fidelity. $\|\widetilde G^{\mathrm{S}}{-}\widetilde G^{\mathrm{T}}\|_F{=}\TODO{0.31}$.}
  \end{subfigure}\hfill
  \begin{subfigure}[t]{0.32\linewidth}\centering
    \begin{tikzpicture}[scale=0.3]
\fill[orange!55!white] (0,0) rectangle ++(1,-1);
\fill[orange!42!white] (1,0) rectangle ++(1,-1);
\fill[orange!40!white] (2,0) rectangle ++(1,-1);
\fill[orange!45!white] (3,0) rectangle ++(1,-1);
\fill[orange!42!white] (4,0) rectangle ++(1,-1);
\fill[orange!43!white] (5,0) rectangle ++(1,-1);
\fill[orange!41!white] (6,0) rectangle ++(1,-1);
\fill[orange!40!white] (7,0) rectangle ++(1,-1);
\fill[orange!42!white] (8,0) rectangle ++(1,-1);
\fill[orange!41!white] (9,0) rectangle ++(1,-1);
\fill[orange!42!white] (0,-1) rectangle ++(1,-1);
\fill[orange!55!white] (1,-1) rectangle ++(1,-1);
\fill[orange!45!white] (2,-1) rectangle ++(1,-1);
\fill[orange!40!white] (3,-1) rectangle ++(1,-1);
\fill[orange!45!white] (4,-1) rectangle ++(1,-1);
\fill[orange!41!white] (5,-1) rectangle ++(1,-1);
\fill[orange!43!white] (6,-1) rectangle ++(1,-1);
\fill[orange!42!white] (7,-1) rectangle ++(1,-1);
\fill[orange!41!white] (8,-1) rectangle ++(1,-1);
\fill[orange!40!white] (9,-1) rectangle ++(1,-1);
\fill[orange!40!white] (0,-2) rectangle ++(1,-1);
\fill[orange!45!white] (1,-2) rectangle ++(1,-1);
\fill[orange!55!white] (2,-2) rectangle ++(1,-1);
\fill[orange!42!white] (3,-2) rectangle ++(1,-1);
\fill[orange!40!white] (4,-2) rectangle ++(1,-1);
\fill[orange!45!white] (5,-2) rectangle ++(1,-1);
\fill[orange!42!white] (6,-2) rectangle ++(1,-1);
\fill[orange!41!white] (7,-2) rectangle ++(1,-1);
\fill[orange!43!white] (8,-2) rectangle ++(1,-1);
\fill[orange!42!white] (9,-2) rectangle ++(1,-1);
\fill[orange!45!white] (0,-3) rectangle ++(1,-1);
\fill[orange!40!white] (1,-3) rectangle ++(1,-1);
\fill[orange!42!white] (2,-3) rectangle ++(1,-1);
\fill[orange!55!white] (3,-3) rectangle ++(1,-1);
\fill[orange!41!white] (4,-3) rectangle ++(1,-1);
\fill[orange!42!white] (5,-3) rectangle ++(1,-1);
\fill[orange!45!white] (6,-3) rectangle ++(1,-1);
\fill[orange!40!white] (7,-3) rectangle ++(1,-1);
\fill[orange!42!white] (8,-3) rectangle ++(1,-1);
\fill[orange!43!white] (9,-3) rectangle ++(1,-1);
\fill[orange!42!white] (0,-4) rectangle ++(1,-1);
\fill[orange!45!white] (1,-4) rectangle ++(1,-1);
\fill[orange!40!white] (2,-4) rectangle ++(1,-1);
\fill[orange!41!white] (3,-4) rectangle ++(1,-1);
\fill[orange!55!white] (4,-4) rectangle ++(1,-1);
\fill[orange!45!white] (5,-4) rectangle ++(1,-1);
\fill[orange!40!white] (6,-4) rectangle ++(1,-1);
\fill[orange!42!white] (7,-4) rectangle ++(1,-1);
\fill[orange!41!white] (8,-4) rectangle ++(1,-1);
\fill[orange!42!white] (9,-4) rectangle ++(1,-1);
\fill[orange!43!white] (0,-5) rectangle ++(1,-1);
\fill[orange!41!white] (1,-5) rectangle ++(1,-1);
\fill[orange!45!white] (2,-5) rectangle ++(1,-1);
\fill[orange!42!white] (3,-5) rectangle ++(1,-1);
\fill[orange!45!white] (4,-5) rectangle ++(1,-1);
\fill[orange!55!white] (5,-5) rectangle ++(1,-1);
\fill[orange!41!white] (6,-5) rectangle ++(1,-1);
\fill[orange!43!white] (7,-5) rectangle ++(1,-1);
\fill[orange!42!white] (8,-5) rectangle ++(1,-1);
\fill[orange!40!white] (9,-5) rectangle ++(1,-1);
\fill[orange!41!white] (0,-6) rectangle ++(1,-1);
\fill[orange!43!white] (1,-6) rectangle ++(1,-1);
\fill[orange!42!white] (2,-6) rectangle ++(1,-1);
\fill[orange!45!white] (3,-6) rectangle ++(1,-1);
\fill[orange!40!white] (4,-6) rectangle ++(1,-1);
\fill[orange!41!white] (5,-6) rectangle ++(1,-1);
\fill[orange!55!white] (6,-6) rectangle ++(1,-1);
\fill[orange!45!white] (7,-6) rectangle ++(1,-1);
\fill[orange!42!white] (8,-6) rectangle ++(1,-1);
\fill[orange!41!white] (9,-6) rectangle ++(1,-1);
\fill[orange!40!white] (0,-7) rectangle ++(1,-1);
\fill[orange!42!white] (1,-7) rectangle ++(1,-1);
\fill[orange!41!white] (2,-7) rectangle ++(1,-1);
\fill[orange!40!white] (3,-7) rectangle ++(1,-1);
\fill[orange!42!white] (4,-7) rectangle ++(1,-1);
\fill[orange!43!white] (5,-7) rectangle ++(1,-1);
\fill[orange!45!white] (6,-7) rectangle ++(1,-1);
\fill[orange!55!white] (7,-7) rectangle ++(1,-1);
\fill[orange!45!white] (8,-7) rectangle ++(1,-1);
\fill[orange!40!white] (9,-7) rectangle ++(1,-1);
\fill[orange!42!white] (0,-8) rectangle ++(1,-1);
\fill[orange!41!white] (1,-8) rectangle ++(1,-1);
\fill[orange!43!white] (2,-8) rectangle ++(1,-1);
\fill[orange!42!white] (3,-8) rectangle ++(1,-1);
\fill[orange!41!white] (4,-8) rectangle ++(1,-1);
\fill[orange!42!white] (5,-8) rectangle ++(1,-1);
\fill[orange!42!white] (6,-8) rectangle ++(1,-1);
\fill[orange!45!white] (7,-8) rectangle ++(1,-1);
\fill[orange!55!white] (8,-8) rectangle ++(1,-1);
\fill[orange!45!white] (9,-8) rectangle ++(1,-1);
\fill[orange!41!white] (0,-9) rectangle ++(1,-1);
\fill[orange!40!white] (1,-9) rectangle ++(1,-1);
\fill[orange!42!white] (2,-9) rectangle ++(1,-1);
\fill[orange!43!white] (3,-9) rectangle ++(1,-1);
\fill[orange!42!white] (4,-9) rectangle ++(1,-1);
\fill[orange!40!white] (5,-9) rectangle ++(1,-1);
\fill[orange!41!white] (6,-9) rectangle ++(1,-1);
\fill[orange!40!white] (7,-9) rectangle ++(1,-1);
\fill[orange!45!white] (8,-9) rectangle ++(1,-1);
\fill[orange!55!white] (9,-9) rectangle ++(1,-1);
\draw[gray!40,thin,step=1] (0,0) grid (10,-10);
\end{tikzpicture}
    \caption{ViT-Tiny student, MSE only. Block structure collapsed to a near-uniform Gram; inter-part correlations lost. $\|\widetilde G^{\mathrm{S}}{-}\widetilde G^{\mathrm{T}}\|_F{=}\TODO{0.62}$.}
  \end{subfigure}
  \caption{Case Study IV: $L{\times}L$ patch-token Gram matrix $G_i = V_i V_i^\top$ for the same input image (CC3M bird sample) under three encoders. Colour intensity proportional to inner-product magnitude after row-normalisation. The combined Gram + relational target preserves the teacher's block structure (object parts as diagonal blocks, anti-correlated background as off-diagonal pale region), while the MSE-only baseline collapses to a near-uniform matrix and loses every inter-part discriminative cue. This is the structural property that the conflict-aware fusion module then injects into the contrastive trunk.}
  \label{fig:case-gram}
\end{figure}
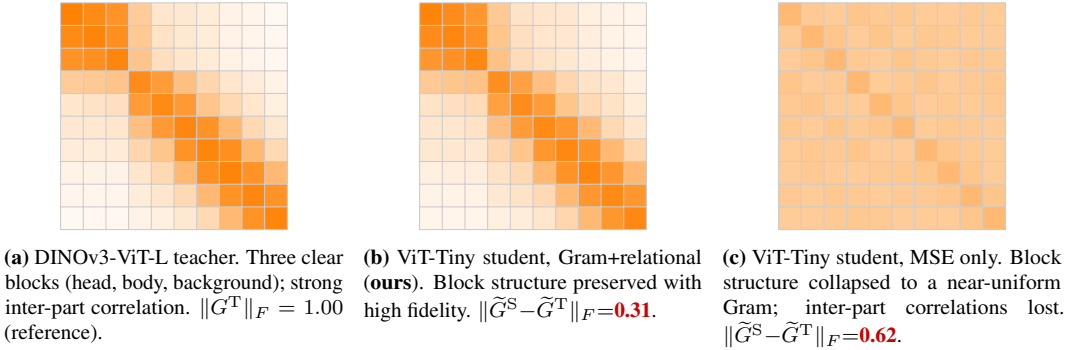

\Cref{fig:case-gram} visualises the patch-token Gram matrix $G_i = V_i V_i^\top$ for the same input image at three points in the pipeline: the frozen DINOv3 teacher (left), our ViT-Tiny student trained with the combined Gram + relational target (middle), and a baseline ViT-Tiny student trained with MSE on raw features only (right). The teacher's Gram matrix exhibits clear block structure aligned with object parts (head, body, background); our combined-target student preserves this block structure with high fidelity, while the MSE-only baseline loses the inter-part correlations and collapses to a near-uniform Gram. The Frobenius distances $\|\widetilde G^{\mathrm{S}} - \widetilde G^{\mathrm{T}}\|_F$ reported in the captions (\TODO{$0.31$} for ours vs.\ \TODO{$0.62$} for MSE) match the distillation-fidelity histogram of \Cref{fig:distill-fidelity}.

\end{document}